\documentclass[lettersize,journal]{IEEEtran}
\usepackage{amsmath,amsfonts}
\usepackage{array}
\usepackage[caption=false,font=normalsize,labelfont=sf,textfont=sf]{subfig}
\usepackage{textcomp}
\usepackage{stfloats}
\usepackage{url}
\usepackage{verbatim}
\usepackage{graphicx}
\usepackage{cite}
\usepackage{multirow}
\usepackage{xcolor}
\usepackage{url}
\usepackage{colortbl}
\usepackage{booktabs}
\usepackage{makecell}
\usepackage{siunitx}
\usepackage[linesnumbered,ruled,vlined]{algorithm2e}
\usepackage{microtype}

\usepackage[colorlinks,citecolor=blue]{hyperref}

\begin{document}

\title{CeRLP: A Cross-embodiment Robot Local Planning Framework for Visual Navigation}

\author{Haoyu Xi$^{1,2}$, 
Mingao Tan$^{1,2}$,
Xinming Zhang$^{2}$,
Siwei Cheng$^{2}$,
Shanze Wang$^{2}$,
Yin Gu$^{2}$,
Xiaoyu Shen$^{2}$,
Wei Zhang$^{2*}$%

\thanks{$^{1}$School of Computer Science, Shanghai Jiao Tong University, Shanghai, P. R. China}%
\thanks{$^{2}$College of Information Science and Technology, Eastern Institute of Technology, Ningbo, P. R. China}%
\thanks{*Corresponding author. Email: zhw@eitech.edu.cn.}%
}



\maketitle
\begin{abstract}
Visual navigation for cross-embodiment robots is challenging due to variations in robot and camera configurations, which can lead to the failure of navigation tasks. Previous approaches typically rely on collecting massive datasets across different robots, which is highly data-intensive, or fine-tuning models, which is time-consuming. Furthermore, both methods often lack explicit consideration of robot geometry. In this paper, we propose a Cross-embodiment Robot Local Planning (CeRLP) framework for general visual navigation, which abstracts visual information into a unified geometric formulation and applies to heterogeneous robots with varying physical dimensions, camera parameters, and camera types. CeRLP introduces a depth estimation scale correction method that utilizes offline pre-calibration to resolve the scale ambiguity of monocular depth estimation, thereby recovering precise metric depth images. Furthermore, CeRLP designs a visual-to-scan abstraction module that projects varying visual inputs into height-adaptive laser scans,  making the policy robust to heterogeneous robots. Experiments in simulation environments demonstrate that CeRLP outperforms comparative methods, validating its robust obstacle avoidance capabilities as a local planner. Additionally, extensive real-world experiments verify the effectiveness of CeRLP in tasks such as point-to-point navigation and vision-language navigation, demonstrating its generalization across varying robot and camera configurations.
\end{abstract}


\section{Introduction}

\IEEEPARstart{M}{obile} robot navigation aims to identify a feasible path to a destination within cluttered environments while ensuring platform safety. This fundamental capability holds immense value for diverse applications, ranging from autonomous logistics to search and rescue operations.
While traditional approaches like the Dynamic Window Approach (DWA)\cite{580977} and Timed Elastic Band (TEB)\cite{6309484, 8206458,10802811} have proven effective, they rely  on precise, pre-built maps for path planning and obstacle avoidance.
Deep Reinforcement Learning (DRL) approaches directly process raw sensor observations from cameras or LiDARs and predict future actions. This paradigm enables robots to learn policies through interaction with the environment without explicit map dependence.

\begin{figure*}[!t]
	\centering
	\includegraphics[width=0.7\textwidth]{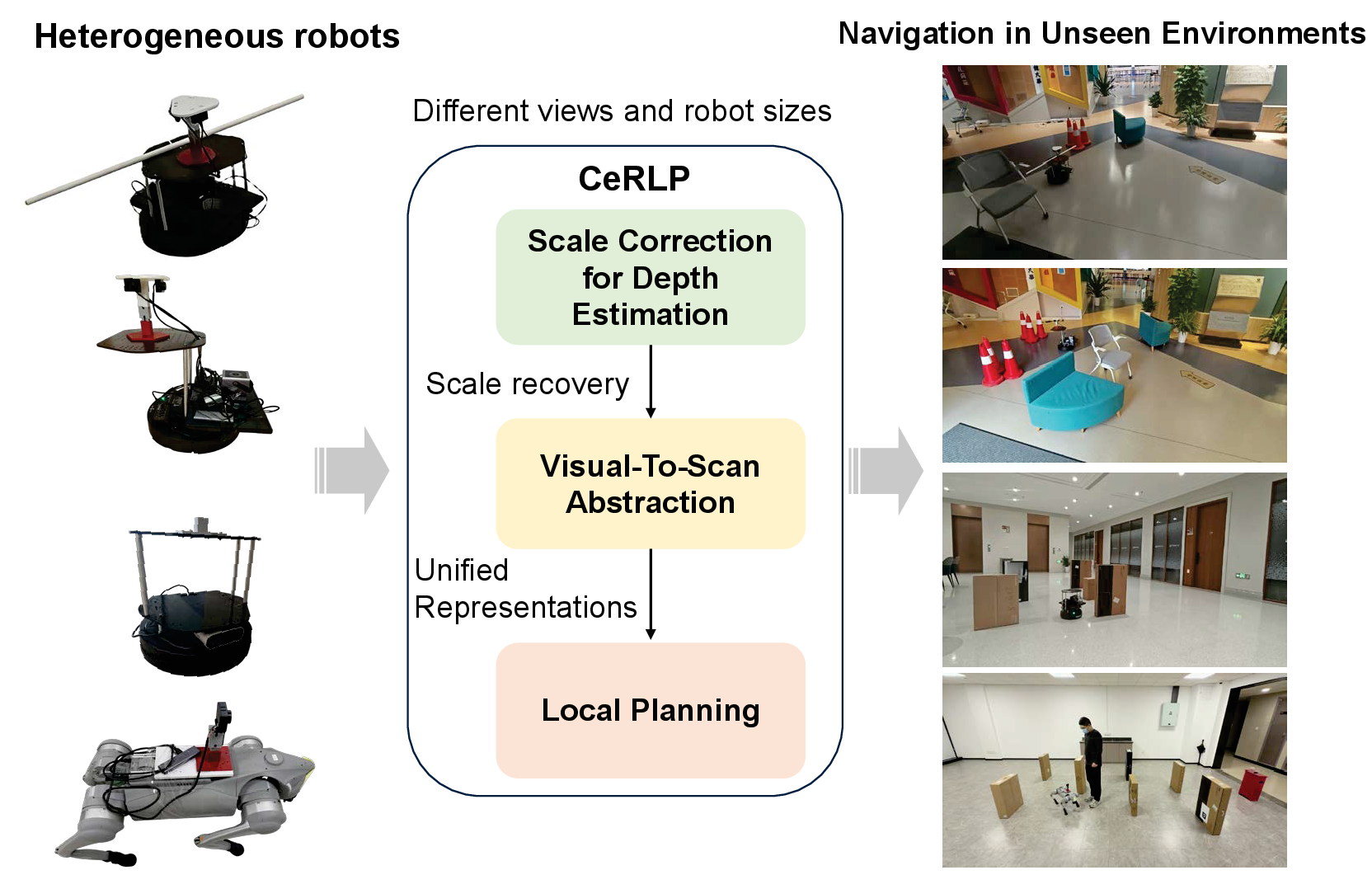}\\
	\caption{ CeRLP addresses the challenge of visual navigation across heterogeneous  robots with varying  physical dimensions and camera configurations. By  transforming diverse visual inputs into a unified  virtual laser  scan and modeling the robot as a generalized cuboid, the framework effectively achieves  obstacle avoidance in unseen environments without  any fine-tuning.
	}\label{fig:fig1}
\end{figure*}

LiDAR provides reliable and accurate ranging information of the surrounding environment\cite{zhu2022analyze}. Prior works\cite{zhang2022ipaprec,zhang2025learning,10089196,10900448} utilize LiDAR to ensure the safety of robot navigation.
Additionally, DRL-DCLP\cite{10900448} has demonstrated that geometric information from laser scans facilitates robust navigation policies transferable across robots with varying physical dimensions. 
Different from LiDARs, cameras offer high frame rates and rich texture information\cite{liu2023real} while typically being cheaper\cite{yasuda2020autonomous}.  Visual navigation has thus seen rapid development, particularly in tasks requiring semantic understanding\cite{10336033,10610025,10801626}.
However, achieving robust generalization across heterogeneous robots remains a challenge in visual navigation, especially for monocular systems. Unlike laser scans which provide normalized metric measurements, raw visual observations suffer from scale ambiguity problem: scale cannot be captured in a single image, as scale information is lost in the process of projecting from a 3D scene to a 2D image.
Consequently, a policy learned on one dataset often experiences a domain shift when deployed on another dataset with a different camera setup\cite{zeng2024rsa}. Even minor alterations in camera mounting height or pitch can lead to a degradation in navigation accuracy, as the agent may misinterpret the distance to obstacles due to the lack of explicit scale cues.
This generalization gap becomes challenging when integrating navigation with high-level cognitive tasks, such as Vision-Language Navigation (VLN)\cite{8578485}. Existing VLN methods primarily focus on grounding language instructions to visual topological graphs, often overlooking the performance of local planning in the physical world. Most VLN agents treat the robot as a simplified point mass, ignoring the risk of obstacle collision in local planning. 

To address these challenges, recent learning-based visual navigation approaches have focused on large-scale data or fine-tuning. ExAug\cite{10160761}, GNM\cite{10161227}, ViNT\cite{pmlr-v229-shah23a}, RING\cite{eftekharone} utilize massive, diverse datasets to learn generalizable visual representations.
Alternatively, adaptation-based methods such as SplitNet\cite{gordon2019splitnet}, FastRLAP\cite{stachowicz2023fastrlap}, OVRL\cite{yadav2023offline} employ fine-tuning to adapt agents to new environments or embodiments.  However, none of these methods explicitly incorporate the robot's geometric information, thereby introducing potential safety risks.
Distinguishing our work from these approaches, we introduce a novel perspective by abstracting visual information into a unified geometric formulation, thereby avoiding the need to directly learn camera-specific pixel distributions.
By representing different visual observations into a standardized virtual laser scan and modeling the robot as a generalized cuboid,  we effectively decouple the navigation policy from specific sensor hardware and robot platform.
Under this formulation, visual navigation becomes agnostic to robot morphology. Wheeled robots, quadrupeds, and other platforms can be uniformly represented as cuboids with varying lengths and widths, equipped with cameras following a consistent scaling convention.

Therefore, this paper proposes CeRLP, a Cross-embodiment Robot Local Planning framework for general visual navigation, as shown in Fig. \ref{fig:fig1}.
\textbf{To the best of our knowledge, CeRLP is the first visual-based local planning framework that enables cross-embodiment deployment across diverse robots without requiring large-scale multi-robot datasets, fine-tuning, or retraining.}
CeRLP applies to varying types of cameras as well as cameras with different intrinsic and extrinsic parameters without any fine-tuning. Meanwhile, CeRLP does not require large and various datasets and is trained only in simulation environments, enabling direct zero-shot transfer to the real world and navigation in unseen environments.
Instead of processing raw RGB images directly, CeRLP utilizes a depth estimation module to obtain relative depth images, which are utilized to recover metric scale and geometric structure.
To handle the scale ambiguity of monocular and camera perspective shifts, CeRLP introduces a scale correction method and a visual-to-scan abstraction module, which transforms relative depth images into metric depth images and obtains the height-adaptive laser scan, effectively eliminating the impact of camera variations.
In summary, the main contributions of this paper are:
\begin{itemize}
\item  We propose a Cross-embodiment Robot Local Planning (CeRLP) framework  for general visual navigation that achieves zero-shot transfer across heterogeneous robots with varying physical dimensions and camera configurations.
    
\item We introduce the scale correction method for monocular depth estimation that utilizes the offline scale calibration to achieve scale recovery, addressing the gap between relative depth images and metric depth images.
    
\item We design the visual-to-scan abstraction method that converts depth information into a virtual laser scan with adaptive height adjustment, making the policy robust to heterogeneous robots.
    
\item We conduct extensive real-world experiments on multiple robot platforms with different camera setups, demonstrating that our method  outperforms baseline approaches in terms of success rate and generalization capability.
\end{itemize}

\section{Related Work}

\subsection{Learning-based Visual Navigation}
Learning-based visual navigation has achieved promising progress in recent years. HALO\cite{seneviratne2025halo} uses an offline reward-learning pipeline for egocentric RGB navigation, and  generalizes to unseen environments and hardware setups not present in the training data. FastRLAP\cite{stachowicz2023fastrlap} takes RGB images as input and trains in the real world without requiring human intervention. However, these approaches  typically lack specific designs for heterogeneous cameras and face inflexibility when robot configurations change.
Different from these, some works have focused on general navigation models that aim to control diverse robots with a single model. GNM\cite{10161227}, ViNT\cite{pmlr-v229-shah23a}, and NoMaD\cite{10610665} leverage large-scale heterogeneous datasets collected from different robots to learn a goal-conditioned policy, which can be deployed on various robots and in different environments.
Nevertheless, these data-driven approaches learn robot capabilities from trajectories rather than explicitly modeling the robot's physical dimensions. Consequently, they cannot guarantee safety in narrow environments when deployed on unseen robots with different footprints. Unlike these data-driven methods, our framework explicitly integrates robot dimension configurations into the visual navigation pipeline, enabling zero-shot transfer to new embodiments without requiring massive heterogeneous datasets.

\subsection{Monocular Depth Estimation and Scale Recovery}

Monocular depth estimation estimates depth from a single RGB image\cite{9750985}, which is a fundamental component for obstacle avoidance in low-cost robots. Recent foundation models \cite{10657693,37379163738604} have demonstrated strong zero-shot generalization capabilities across diverse scenes.
However, these models typically predict relative depth, which suffers from scale ambiguity, making it unsuitable for direct metric navigation and collision avoidance.
To recover metric scale, several approaches have been proposed.
Louw et al. \cite{louw2025practical} manually set the scale factor applied to specific cameras, but this method lacks generalizability. 
RSA\cite{zeng2024rsa} utilizes language descriptions to align relative predictions with metric scales, while HybridDepth\cite{ganj2025hybriddepth} leverages focal stack information to provide reliable metric depth for scale prior calculation. 
Metric3D \cite{10378288} proposes resolving scale ambiguity by transforming input images into a canonical camera space. UniDepth \cite{piccinelli2024unidepth} explicitly estimates camera intrinsics to recover metric scale. VGGT\cite{11094896} introduces a visual geometry grounded transformer that jointly infers camera parameters and 3D structure in a feed-forward manner.
However, these methods still have errors when deployed on specific camera setups, directly affecting safety in obstacle avoidance tasks.
In continuous navigation tasks, pure geometric recovery methods often face numerical instability, leading to ill-conditioned matrices when object depths are uniform.
This paper addresses the scale ambiguity problem by proposing a  scale correction module. Instead of retraining depth models, we formulate scale recovery as an optimization problem constrained by ground-plane geometry, ensuring robust metric depth estimation for safe navigation.

\subsection{Cross-Embodiment and Dimension-Configurable Planning}

Adapting navigation policies to different robot morphologies is a critical challenge.
Traditional approaches often rely on manual parameter tuning for planners like DWA\cite{580977} or TEB\cite{6309484, 8206458,10802811}. Recent learning-based works attempt to automate this; for instance, Lu et al.\cite{10749994} use demonstration trajectories to guide parameter tuning for RL.
However, visual navigation faces the additional challenge of sensor variation. Changes in camera height, pitch, or field of view (FOV) cause severe distribution shifts in observations. ExAug\cite{10160761} addresses this by training with a view augmentation module that projects observations from different platforms into a unified synthetic view. 
However, this method still requires large datasets for training and assumes the robot shape is cylindrical.
X-Nav\cite{11244111} takes depth image as input and uses knowledge distillation to train a general navigation policy for a variety of wheeled and quadrupedal robots.
For LiDAR-based navigation, DRL-DCLP\cite{10900448} successfully achieves zero-shot adaptation to rectangular differential-drive robots with varying sizes. However, a counterpart for visual navigation that simultaneously handles visual sensor variations and physical body variations is still lacking.
Our work bridges this gap by proposing a unified, dimension-configurable visual navigation framework. Our method handles diverse sensor positions and robot dimensions in a zero-shot manner, eliminating the need for post-fine-tuning or platform-specific data collection.

\section{Preliminaries}
In this section, we first define the coordinate systems, formulate the cross-embodiment visual navigation problem, and analyze the scale ambiguity in monocular depth estimation. Finally, we briefly review the baseline local planning method.

\subsection{Coordinate Systems}
Firstly, we define three coordinate systems essential for our framework:
\begin{itemize}
    \item \textbf{World Frame ($\mathcal{W}$):} The global frame where the robot navigation task is defined.
    \item \textbf{Robot Frame ($\mathcal{R}$):} The local frame attached to the robot base, with the y-axis pointing forward and the x-axis pointing right. The robot is modeled as a cuboid with dimensions $\mathbf{body} = [L_{\text{front}}, L_{\text{rear}}, W]$. $L_{\text{front}}$ and $L_{\text{rear}}$ represent the distances from the drive center to the front and rear, respectively. $W$ is the width of the robot.
    \item \textbf{Camera Frame ($\mathcal{C}$):} The sensor frame attached to the camera. The transformation from $\mathcal{C}$ to $\mathcal{R}$ is defined by extrinsic parameters $\mathbf{T}_{\text{ext}} = [\mathbf{R}_{\text{ext}} | \mathbf{t}_{\text{ext}}]$, where $\mathbf{t}_{\text{ext}}$ includes the mounting height, and $\mathbf{R}_{\text{ext}}$ accounts for the pitch angle. 
    The camera is fixed relative to the robot, where $x_{\text{cam}}$ and $y_{\text{cam}}$ indicate the relative distance to the robot's drive center, and $z_{\text{cam}}$ represents the camera's height above the ground.

\end{itemize}

\subsection{Problem Formulation}
We define the cross-embodiment visual navigation task as a continuous control problem where a robot must navigate to a target position within an unknown environment without collision.

\subsubsection{Embodiment Representation}
To handle diverse robot platforms, we explicitly model the physical properties of the agent. An embodiment $e \in \mathcal{E}$ is characterized by the tuple $e = (\mathbf{body}, \mathbf{K}, \mathbf{T}_{\text{ext}}, \mathbf{L}_{\text{dyn}})$. Here, $\mathbf{body} = [L_{\text{front}}, L_{\text{rear}}, W]$ represents the physical dimensions, $\mathbf{K}$ and $\mathbf{T}_{\text{ext}}$ denote the camera intrinsic and extrinsic parameters, and $\mathbf{L}_{\text{dyn}}$ defines the velocity and acceleration limits.

\subsubsection{Observation and Action Space}
At each time step $t$, the robot receives a raw visual observation $\mathbf{I}_{\text{t}} \in \mathbb{R}^{H \times W \times 3}$ from the onboard camera and a relative goal position $\mathbf{g}_{\text{t}} \in \mathbb{R}^2$. The objective is to compute a continuous velocity command $\mathbf{u}_{\text{t}} = (v_{\text{t}}, \omega_{\text{t}})$ to drive the robot towards the goal.

\subsubsection{Unified Geometric Representation}
A fundamental challenge in this setting is the coupling between the raw visual observation $\mathbf{I}_{\text{t}}$ and the embodiment-specific camera parameters $(\mathbf{K}, \mathbf{T}_{\text{ext}})$. Direct learning from $\mathbf{I}_{\text{t}}$ often leads to poor generalization across different cameras. To address this, we propose a mapping function $\Phi: (\mathbf{I}_{\text{t}}, \mathbf{K}, \mathbf{T}_{\text{ext}}) \rightarrow \mathbf{xr}_{\text{t}}$, which transforms the heterogeneous visual inputs into a unified geometric representation $\mathbf{xr}_{\text{t}}$. In this work, $\mathbf{xr}_{\text{t}}$ takes the form of a unified virtual laser scan. This representation is invariant to specific camera configurations, providing a consistent state space for the planner.

\subsection{Scale Ambiguity in Monocular Depth Estimation}

Monocular depth estimation aims to map a single RGB image $I$ to a depth image $D$. However, strictly recovering absolute metric depth from a single image is ill-posed due to the projective nature of cameras. The ambiguity arises from two primary sources: scale ambiguity and shift ambiguity.
Deep learning-based monocular depth estimation models\cite{555529690332969091,9178977,10657693,37379163738604} are typically trained on datasets mixing various sources. To ensure robust convergence across diverse datasets with conflicting scales, these models are trained in a scale-invariant or affine-invariant space. Specifically, the prediction is often aligned to zero translation and unit scale during training. Consequently, the raw output $d_{\text{pred}}$ is the inverse relative depth $d_{\text{gt}} = 1/Z_{\text{gt}}$.

Based on the perspective in~\cite{zeng2024rsa, 9178977}, the relationship between the ground-truth inverse depth $d_{\text{gt}}$ and the model prediction $d_{\text{pred}}$ can be formulated as a linear transformation in the disparity domain:
\begin{equation}
d_{\text{gt}} = s_{\text{1}} \cdot d_{\text{pred}} + s_{\text{2}},
\end{equation}
where $s_{\text{1}} \in \mathbb{R}^+$ represents the global scale factor, and $s_{\text{2}} \in \mathbb{R}$ represents the disparity shift. 
The $s_{\text{1}}$ accounts for the unknown physical baseline in the single image, while the $s_{\text{2}}$ compensates for the difference in the definition of the depth range.

Therefore, to deploy a generic monocular depth model for robotic navigation, it is necessary to identify the optimal parameters $(s_{\text{1}}, s_{\text{2}})$ that project the relative depth into the metric depth.

\subsection{Dimension-Configurable Planning}
The CeRLP local planning module, built on DRL-DCLP \cite{10900448}, addresses local navigation as a deep reinforcement learning (DRL) problem. It generates real-time velocity commands \(\mathbf{u}_{\text{t}} = (v_{\text{t}}, \omega_{\text{t}})\) for target-oriented obstacle avoidance using only local observations. To achieve generalization across diverse robotic platforms, the method integrates a dimension-configurable policy network with a curriculum learning framework.

At each time step $t$, the policy input $\mathbf{o}_{\text{t}}$ aggregates the unified geometric representation and the robot state:
\begin{equation}\label{eq:observation}
\mathbf{o}_{\text{t}} = \{ \mathbf{xr}_{\text{t}},\; \mathbf{g}_{\text{t}},\; \mathbf{v}_{\text{t}},\; \mathbf{L}_{\text{dyn}} \},
\end{equation}
where $\mathbf{v}_{\text{t}} = (v_{\text{t}}, \omega_{\text{t}})$ denotes the current velocity, and $\mathbf{L}_{\text{dyn}}$ includes velocity limits $(v_{\text{max}}, \omega_{\text{max}})$ and acceleration limits ($a_{v, \text{max}}, a_{\omega, \text{max}}$).

Crucially, the unified geometric representation $\mathbf{xr}_{\text{t}}$ is processed as a set of points $\mathbf{xr}_{\text{t}} = \{\mathbf{p}_{\text{t},\text{i}}\}_{\text{i}=1}^{\text{n}}$. To explicitly incorporate the robot's geometry into the decision-making process, each point is encoded with dimension-aware features:
\begin{equation}\label{eq:point_feature}
\mathbf{p}_{\text{t},\text{i}} = \left( \sin\phi_{\text{t},\text{i}},\; \cos\phi_{\text{t},\text{i}},\; \frac{1}{d_{\text{t},\text{i}}-\beta},\; L_{\text{front}},\; L_{\text{rear}},\; \frac{W}{2} \right),
\end{equation}
where $d_{\text{t},\text{i}}$ and $\phi_{\text{t},\text{i}}$ are the distance and angle of the $i$-th scan point, and $\beta$ is a trainable scaling parameter \cite{9804689}. This formulation directly embeds the robot's physical dimensions $\mathbf{body} = [L_{\text{front}}, L_{\text{rear}}, W]$ into the perception features. The point set is processed by a PointNet encoder \cite{Qi2016PointNetDL} to extract a global geometric feature information, which is then concatenated with the goal and state information. The final action is generated by a Soft Actor-Critic (SAC) \cite{haarnoja18b} network.

We employ a curriculum learning strategy in ROS Stage simulation\cite{vaughan2008massively} to train the policy over a continuous space of robot configurations. By progressively increasing the complexity of the environment and the diversity of the sampled physical dimensions parameters $\mathbf{body}$, the policy learns to adapt its obstacle avoidance behavior to the specific physical constraints of the robot.

\section{Methods}
\subsection{Overall framework of CeRLP}

\begin{figure*}[!t]
	\centering
	\includegraphics[width=\textwidth]{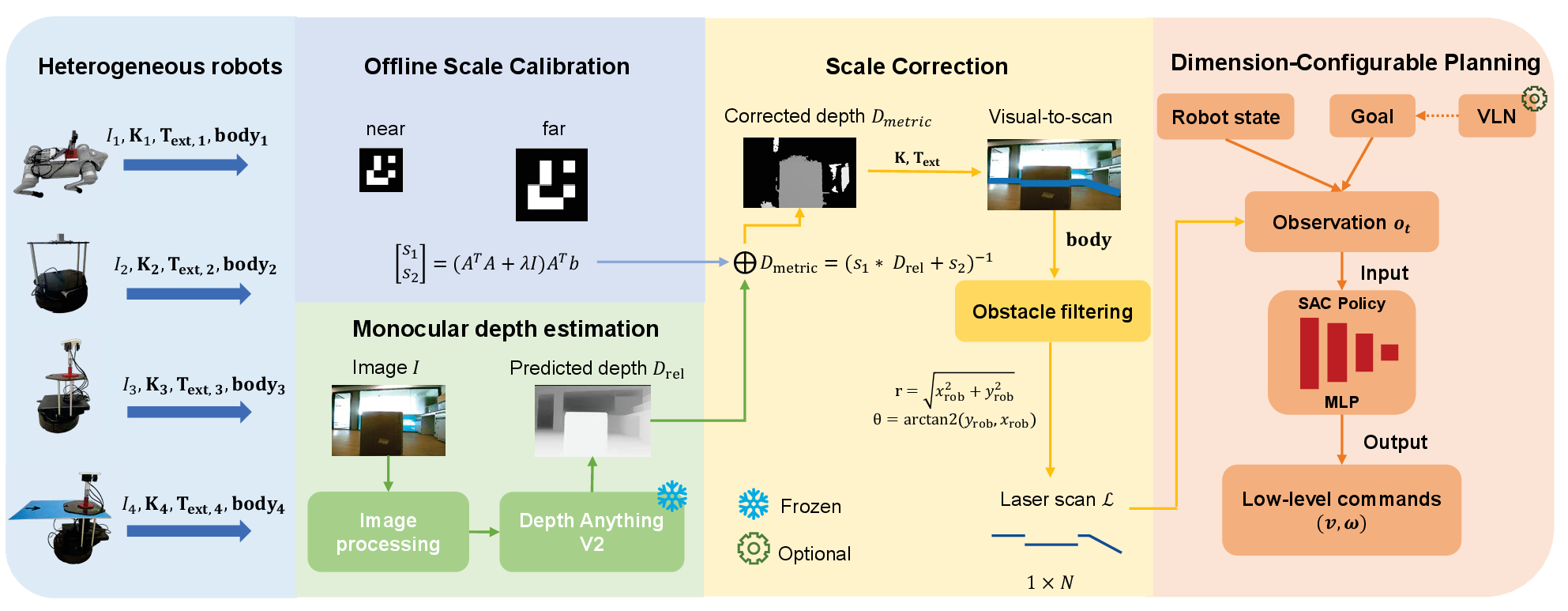}\\
	\caption{ The overall framework of CeRLP. The system abstracts visual observations from heterogeneous robots into a unified geometric representation. Heterogeneous robots with differing intrinsic parameters, extrinsic parameters, and physical dimensions capture RGB images from cameras. Scale calibration is an offline process utilizing distant and nearby ArUco markers to calculate the scale factor and disparity shift, thereby achieving scale recovery. Depth estimation employs pre-trained Depth Anything V2 to predict relative depth images. Scale Correction rectifies the relative depth image into a metric depth image during inference, subsequently transforming this metric depth image into a height-adaptive virtual laser scan. The dimension-configurable policy utilises this unified laser scan, goal position $\mathbf{g}_{\text{t}}$, and robot state information to generate safe control commands, achieving zero-shot transfer across embodiments. The robot state information contains physical dimensions $\mathbf{body}$, current velocity $\mathbf{v}_{\text{t}}$, and velocity and acceleration limits $\mathbf{L}_{\text{dyn}}$.  Furthermore, this module can receive high-level commands, functioning as the lower-level obstacle avoidance planner for VLN.
	}\label{fig:fig2}
\end{figure*}

The proposed CeRLP framework aims to bridge the gap between visual perception and geometric navigation constraints. As illustrated in Fig. \ref{fig:fig2},  the pipeline consists of three core modules: (1) Scale Correction for Depth Estimation, which recovers metric scale from monocular images using a calibration-based rectification; (2) Visual-to-Scan Abstraction, which transforms the metric depth into a unified virtual laser scan, explicitly handling camera extrinsic variations; and (3) Dimension-Configurable Planning, which generates velocity commands based on the homogenized observation and robot dimensions. 

\subsection{Scale Correction for Depth Estimation}

\subsubsection{Scale-Aware Depth Estimation}

Recent advancements in foundational vision models, such as Depth Anything~\cite{10657693,37379163738604} and MiDaS~\cite{9178977}, have demonstrated exceptional zero-shot generalization capabilities across diverse scenarios. In this work, we employ a pre-trained depth estimation model, denoted as $\mathcal{F}$, as the backbone to extract geometric features from RGB images. Given an input image $I$, the model predicts a relative depth image ${{D}}_{\text{rel}} = \mathcal{F}(I)$. Before being input to the depth estimation model, the original image is processed by normalizing and standardizing using ImageNet statistics\cite{deng2009imagenet}.

However, these models are typically trained with scale-invariant losses to ensure adaptability to diverse datasets. Consequently, the output ${{D}}_{\text{rel}}$ is defined up to an unknown affine transformation relative to the true metric inverse depth. Although it is sufficient for topological scene understanding, this lack of metric scale renders ${D}_{\text{rel}}$ inadequate for direct collision checking and path planning in physical environments.

\subsubsection{Offline Scale Calibration}

To address the scale ambiguity across different cameras, we first propose an offline scale calibration method. This process utilizes the camera intrinsic matrix and distortion coefficients as inputs. By detecting ArUco markers of known physical sizes, we leverage geometric constraints to estimate the specific scale and shift parameters for the depth image. The problem is formulated as a least-squares optimization, outputting a scale factor and a shift value. The overall pipeline describes the following mapping:

\begin{equation}
f: (I, D_{\text{rel}}) \mapsto (D_{\text{metric}}, s_{\text{1}}, s_{\text{2}}),
\end{equation}
where $D_{\text{metric}}$ denotes the corrected metric depth image, ${D}_{\text{rel}}$ is the predicted relative depth image by depth estimation model, and $s_{\text{1}}$ and $s_{\text{2}}$ are the estimated calibration parameters. 

Our approach establishes an affine transformation between relative and absolute depth, solving for the parameters using known geometric priors to achieve metric recovery.

First, the input RGB image is converted to grayscale to detect ArUco markers from a predefined dictionary. For each detected marker $i$, we extract the pixel coordinates of its four corners, $\mathbf{c}_{\text{i}} = [u_{\text{k}}, v_{\text{k}}]^T, k \in \{1,2,3,4\}$. Given the known physical size $s'$ of the marker, the coordinates of the corners in the object coordinate system, $\mathbf{P}_{\text{obj}}$, are defined as:
\begin{equation}
\mathbf{P}_{\text{obj}} = \left[\begin{array}{rrr}
-s'/2 & s'/2 & 0 \\
s'/2 & s'/2 & 0 \\
s'/2 & -s'/2 & 0 \\
-s'/2 & -s'/2 & 0
\end{array}\right]^T.
\end{equation}

We employ the Perspective-n-Point (PnP) algorithm to estimate the 6-DoF pose of the marker relative to the camera. Using the camera intrinsic matrix $\mathbf{K}$ and distortion coefficients $\mathbf{d}$, we solve for the rotation vector $\mathbf{r}_{\text{vec}}$ and translation vector $\mathbf{t}_{\text{vec}}$:
\begin{equation}
(\mathbf{r}_{\text{vec}}, \mathbf{t}_{\text{vec}}) = \text{solvePnP}(\mathbf{P}_{\text{obj}}, \mathbf{c}, \mathbf{K}, \mathbf{d}).
\end{equation}

Subsequently, the spatial coordinates of the marker corners in the camera coordinate system, $\mathbf{P}_{\text{cam}}$, are computed using the Rodrigues transformation:
\begin{equation}
\begin{split}
\mathbf{R} &= \text{Rodrigues}(\mathbf{r}_{\text{vec}}), \\
\mathbf{P}_{\text{cam}}^{(\text{k})} &= \mathbf{R} \cdot \mathbf{P}_{\text{obj}}^{(\text{k})} + \mathbf{t}_{\text{vec}}.
\end{split}
\end{equation}

The ground truth metric depth $z_{\text{real}}^{(\text{k})}$ for each corner corresponds to the Z-component of $\mathbf{P}_{\text{cam}}^{(\text{k})}$:
\begin{equation}
z_{\text{real}}^{(\text{k})} = [\mathbf{P}_{\text{cam}}^{(\text{k})}]_{\text{z}}.
\end{equation}

With the ground truth depth $z_{\text{real}}$ obtained from ArUco geometry calculation and the predicted relative values $D_{\text{rel}}(u_{\text{p}},v_{\text{p}})$ sampled from the depth model at corresponding pixel coordinates, we construct a linear system $\mathbf{A}\mathbf{x} = \mathbf{b}$ to solve for the parameters:
\begin{equation}
\mathbf{A} = \begin{bmatrix}
D_{\text{rel}}(u_{\text{p,1}}, v_{\text{p,1}}) & 1 \\
D_{\text{rel}}(u_{\text{p,2}}, v_{\text{p,2}}) & 1 \\
\vdots & \vdots \\
D_{\text{rel}}(u_{\text{p,n}}, v_{\text{p,n}}) & 1
\end{bmatrix}, \quad
\mathbf{b} = \begin{bmatrix}
{z_{\text{real}}^{(\text{1})}}^{-1} \\
{z_{\text{real}}^{(\text{2})}}^{-1} \\
\vdots \\
{z_{\text{real}}^{(\text{n})}}^{-1}
\end{bmatrix}, \quad
\mathbf{x} = \begin{bmatrix} s_{\text{1}} \\ s_{\text{2}} \end{bmatrix},
\end{equation}
where $n$ is the total number of corners from detected markers. 

To ensure the robustness of the solution, we utilize markers placed at varying distances to capture sufficient geometric information. Specifically, since the linear system can be ill-conditioned when points are spatially clustered, we employ Ridge Regression\cite{23071271436} to stabilize the estimation.
Furthermore,  we require calibration data from at least a near and a far distances to constrain the affine transformation effectively and prevent overfitting to a single depth plane. 

The parameters are obtained by minimizing the regularized least squares objective:
\begin{equation}
\min_{\mathbf{x}} \left( \| \mathbf{A}\mathbf{x} - \mathbf{b} \|_{\text{2}}^{\text{2}} + \lambda \| \mathbf{x} \|_{\text{2}}^{\text{2}} \right).
\end{equation}

Consequently, the analytical closed-form solution is given by:
\begin{equation}
\mathbf{x} = (\mathbf{A}^T\mathbf{A} + \lambda\mathbf{I})^{-1}\mathbf{A}^T\mathbf{b},
\end{equation}
where $\lambda$ is the regularization parameter and $\mathbf{I}$ denotes the identity matrix. The whole process of offline scale calibration is shown in  Algorithm \ref{alg:calibration}.

\begin{algorithm}[t]
\caption{Offline Scale Calibration}
\label{alg:calibration}
\SetAlgoLined
\DontPrintSemicolon

\KwIn{Calibration Images $I$ $\in$ $I_{\text{calib}}$, Marker Size $s$, Intrinsic $\mathbf{K}$, Distortion $\mathbf{d}$, Depth Model $\mathcal{F}$}
\KwOut{Calibration Parameters $\mathbf{x} = [s_{\text{1}}, s_{\text{2}}]^T$}

Initialize linear system matrices $\mathbf{A} \leftarrow \emptyset, \mathbf{b} \leftarrow \emptyset$\;

\For{ \upshape each image $I$ }{
    \tcp{covering near and far distances}
    Predict relative depth image: $D_{\text{rel}} \leftarrow \mathcal{F}(I)$\;
    Detect ArUco corners $\mathbf{c}$ in $I$\;
    Estimate marker pose $(\mathbf{r}_{\text{vec}}, \mathbf{t}_{\text{vec}})$ via PnP using $\mathbf{K}, \mathbf{d}$\;
    
    \For{\upshape detected corner $j$}{
        Compute ground truth depth $z_{\text{real}}^{(\text{j})}$ from pose transformation\;
        Sample predicted depth $d_{\text{pred}}^{(\text{j})}$ at corner pixel $(u_{\text{p, j}}, v_{\text{p, j}})$\;
        Append row $[d_{\text{pred}}^{(\text{j})}, 1]$ to $\mathbf{A}$\;
        Append value ${z_{\text{real}}^{(\text{j})}}^{-1}$ to $\mathbf{b}$\;
    }
}

Solve optimization problem: $\min_{\mathbf{x}} (\| \mathbf{A}\mathbf{x} - \mathbf{b} \|_{\text{2}}^{\text{2}} + \lambda \| \mathbf{x} \|_{\text{2}}^{\text{2}})$\;
Compute closed-form solution: $\mathbf{x} \leftarrow (\mathbf{A}^T\mathbf{A} + \lambda\mathbf{I})^{-1}\mathbf{A}^T\mathbf{b}$\;

\Return $\mathbf{x}$\;
\end{algorithm}

\subsubsection{Online Scale Correction}
During real-time navigation, the pre-calculated parameters $s_{\text{1}}$ and $s_{\text{2}}$ are applied to the depth estimation model output for inference. For each RGB image $I$, the metric depth image $D_{\text{metric}}$ is generated element-wise:
\begin{equation}
    D_{\text{metric}} = \frac{1}{s_{\text{1}}\cdot  \mathcal{F}(I) + s_{\text{2}}}.
\end{equation}

This rectified depth image $D_{\text{metric}}$ serves as the input for the subsequent Visual-to-Scan abstraction module.

\subsection{Visual-to-Scan Abstraction}

To achieve robust navigation across diverse robot embodiments and camera setups, we introduce the Visual-to-Scan Abstraction module, which converts corrected depth images into a unified 2D laser scan. This process involves reconstructing the 3D world from images and then slicing it horizontally based on the robot's physical size. Due to variations in camera positions, extracting LiDAR data from depth images requires avoiding the low merging intervals, which is sensitive to camera pitch and mounting height. Our approach leverages  3D geometric information, which ensures that the generated laserscan scan strictly adheres to the physical passability constraints of the specific robot.

\subsubsection{3D Back-projection and Transformation}
Given the metric depth image $D_{\text{metric}}$ and the camera intrinsic parameters $\mathbf{K} = \{f_{\text{x}}, f_{\text{y}}, c_{\text{x}}, c_{\text{y}}\}$, we first back-project each valid pixel $(u_{\text{p}},v_{\text{p}})$ with depth $Z(u_{\text{p}},v_{\text{p}})$ into the 3D camera coordinate frame, obtaining 3D Point Cloud in the Robot Coordinate Frame $\mathcal{P}_{\text{rob}}$. The point $\mathbf{p}_{\text{cam}} = [X_{\text{c}}, Y_{\text{c}}, Z_{\text{c}}]^T$ is computed as:

\begin{equation} 
\begin{aligned}
X_{\text{c}} &= \frac{(u_{\text{p}} - c_{\text{x}}) Z(u_{\text{p}},v_{\text{p}})}{f_{\text{x}}}, \\
Y_{\text{c}} &= \frac{(v_{\text{p}}) - c_{\text{y}}) Z(u_{\text{p}},v_{\text{p}})}{f_{\text{y}}}, \\
Z_{\text{c}} &= Z(u_{\text{p}},v_{\text{p}}).
\end{aligned}
\end{equation}

To account for varying sensor placements, we transform these points from the camera frame $\mathcal{C}$ to the robot frame $\mathcal{R}$. Let the camera extrinsic parameters be $\mathbf{T}_{\text{ext}} = [\mathbf{R}_{\text{ext}} | \mathbf{t}_{\text{ext}}]$, where $\mathbf{t}_{\text{ext}}$ includes the mounting height, and $\mathbf{R}_{\text{ext}}$ accounts for the pitch angle. The point in the robot frame $\mathbf{p}_{\text{rob}} = [x_{\text{rob}}, y_{\text{rob}}, z_{\text{rob}}]^T$ is obtained by:
\begin{equation}
\mathbf{p}_{\text{rob}} = \mathbf{R}_{\text{ext}} \mathbf{p}_{\text{cam}} + \mathbf{t}_{\text{ext}}.
\end{equation}

This transformation effectively neutralizes the perspective distortion caused by camera tilt, aligning the ground plane in a unified coordinate system.

\subsubsection{Height-Adaptive Obstacle Filtering}
In the robot frame $\mathcal{R}$, we apply a height-based filter tailored to the specific robot's dimensions. This step decides which points are actual obstacles. It removes the ground and ignores objects hanging much higher than the robot in 3D Point Cloud $\mathcal{P}_{\text{rob}}$, ensuring the robot only reacts to things it might hit. We define a valid obstacle height range $[h_{\text{min}}, h_{\text{max}}]$, where $h_{\text{min}}$ is a threshold slightly above the ground to filter out floor noise, and $h_{\text{max}}$ corresponds to the robot's physical height. The output is filtered obstacle point cloud $\mathcal{P}_{\text{obs}}$.

We generate a set of effective obstacle points $\mathcal{P}_{\text{obs}}$ by retaining only those points within the collision-relevant vertical space:
\begin{equation}
\mathcal{P}_{\text{obs}} = \{ \mathbf{p}_{\text{rob}} \mid h_{\text{min}} < z_{\text{rob}} < h_{\text{max}} \}.
\end{equation}

This step enables zero-shot transferability: a tall robot will detect overhanging obstacles defined by a larger $h_{\text{max}}$, while a low-profile robot will ignore them, naturally adapting the safety logic to the embodiment.

\subsubsection{Virtual Scan Projection}

Finally, we project the filtered 3D obstacle points onto the navigation plane to synthesize the 2D laser scan $\mathcal{L}$. For each point in $\mathcal{P}_{\text{obs}}$, we calculate its polar coordinates relative to the robot center. The distance is $r = \sqrt{x_{\text{rob}}^{\text{2}} + y_{\text{rob}}^{\text{2}}}$ and the angle is $\theta = \arctan2(y_{\text{rob}}, x_{\text{rob}})$.

The field of view is discretized into angular bins (sectors) corresponding to the desired scan resolution. For each bin $k$, the scan range $\mathcal{L}[k]$ is determined by the minimum distance of all points falling within that sector:
\begin{equation}
\mathcal{L}[k] = \min \{ r \mid (r, \theta) \in \mathcal{P}_{\text{obs}}, \theta \in [\theta_{\text{k}}, \theta_{\text{k}+\text{1}}) \}.
\end{equation}

If no points fall into a bin, the value is set to the sensor's maximum range. The complete online inference pipeline is summarized in Algorithm \ref{alg:visual_to_scan}.

\begin{algorithm}[t]
\caption{Online Scale Correction and Visual-to-Scan Abstraction}
\label{alg:visual_to_scan}
\SetAlgoLined
\DontPrintSemicolon

\KwIn{RGB Image $I$, Calib Params $[s_{\text{1}}, s_{\text{2}}]$, Intrinsic $\mathbf{K}$, Extrinsic $\mathbf{T}_{\text{ext}}$, Height Limits $[h_{\text{min}}, h_{\text{max}}]$}
\KwOut{Virtual Laser Scan $\mathcal{L}$}

\tcp{Step 1: Scale-Aware Depth Recovery}
Predict relative depth: $D_{\text{rel}} \leftarrow \mathcal{F}(I)$\;
Recover metric depth: $D_{\text{metric}} \leftarrow (s_{\text{1}} \cdot D_{\text{rel}} + s_{\text{2}})^{-1}$\;

\tcp{Step 2: Height-Adaptive Point Cloud Generation}
Initialize obstacle set $\mathcal{P}_{\text{obs}} \leftarrow \emptyset$\;
\For{\upshape valid pixel $(\mathrm{u}_{\text{p}},\mathrm{v}_{\text{p}})$ in $\mathrm{D}_{\text{metric}}$}{
    $Z \leftarrow D_{\text{metric}}(u_{\text{p}},v_{\text{p}})$\;
    $\mathbf{p}_{\text{cam}} \leftarrow [ \frac{(u_{\text{p}} - c_{\text{x}})Z}{f_{\text{x}}}, \frac{(v_{\text{p}} - c_{\text{y}})Z}{f_{\text{y}}}, Z ]^T$\;
    $\mathbf{p}_{\text{rob}} \leftarrow \mathbf{R}_{\text{ext}} \mathbf{p}_{\text{cam}} + \mathbf{t}_{\text{ext}}$\;
    
    \If{\upshape $h_{\text{min}} < \mathbf{p}_{\text{rob}} (z) < h_{\text{max}}$}{
        Add $\mathbf{p}_{\text{rob}}$ to $\mathcal{P}_{\text{obs}}$\;
    }
}

\tcp{Step 3: Virtual Scan Projection}
Initialize scan array $\mathcal{L}$ with max sensor range\;
\For{\upshape point $(\mathrm{x}_{\text{rob}}, \mathrm{y}_{\text{rob}}, \mathrm{z}_{\text{rob}})$ in $\mathcal{P}_{\text{obs}}$}{
    Convert to polar coords: $r \leftarrow \sqrt{{x_{\text{rob}}}^{\text{2}} + {y_{\text{rob}}}^{\text{2}}}, \quad \theta \leftarrow \arctan2(y_{\text{rob}}, x_{\text{rob}})$\;
    Determine angular bin index $k$ for $\theta$\;
    Update bin with minimum distance: $\mathcal{L}[k] \leftarrow \min(\mathcal{L}[k], r)$\;
}

\Return $\mathcal{L}$\;
\end{algorithm}

\subsection{Planning}

Following the abstraction of visual data into a unified laser scan, we employ the dimension-configurable policy built on DRL-DCLP\cite{10900448} as the CeRLP local planner.
The details of DRL-DCLP are described in Sec. III-D.
The inputs of CeRLP local planner contain laser scan $\mathcal{L}$ generated by visual-to-scan abstraction and the robot state information including physical dimensions $\mathbf{body}$, relative goal position $\mathbf{g}_{\text{t}}$, current velocity $\mathbf{v}_{\text{t}}$, and velocity and acceleration limits $\mathbf{L}_{\text{dyn}}$.

Besides point-to-point tasks, Vision-Language Navigation (VLN) module is optional to serve as a high-level planner for CeRLP.
We employ a brief VLN framework and our key objective is to validate CeRLP as a lower-level planner to enhance obstacle avoidance capabilities for VLN. CLIP~\cite{pmlr-v139-radford21a} is used to provide high-level navigation commands by processing input RGB images. The camera image is divided into three regions: left, center, and right.  We use CLIP to compute the similarity confidence between each image region and the input language instruction.  
For each region, we utilize CLIP to compute a confidence score, representing the probability that the target object appears in that direction. We identify the optimal direction $re^*$ with the maximum confidence $S_{\text{max}}$.
Based on these confidence scores, the VLN module generates high-level commands in the form of $(d_{\text{cmd}}, \theta_{\text{cmd}})$ pairs, where $d_{\text{cmd}}$ is the forward distance, and the heading angle $\theta_{\text{cmd}}$ corresponds to the direction of the region with the highest confidence. 
The navigation command is derived directly from $S_{\text{max}}$ and the optimal angle $\phi_{re^*}$ using a piecewise function that prioritizes exploration when confidence is low and accelerates approach when confidence is high:

\begin{equation}
    \theta_{\text{cmd}} = 
    \begin{cases} 
        \phi_{\mathrm{re}^*} & \text{if } S_{\text{max}} > 0.65 \\
        0 & \text{otherwise}
    \end{cases},
\end{equation}

\begin{equation}
    d_{\text{cmd}} = 
    \begin{cases} 
        1.0 + 0.3 \cdot S_{\text{max}} & \text{if } S_{\text{max}} > 0.80 \\
        2.0 + 0.5 \cdot S_{\text{max}} & \text{if } 0.65 < S_{\text{max}} \le 0.80 \\
        3.0 & \text{otherwise}
    \end{cases}.
\end{equation}

Finally, these high-level commands are  transformed into the world frame for the lower-level planner, which avoids obstacles and ensures safe execution of high-level instructions. If the optimal direction is center for several consecutive frames and exceeds the threshold, the destination is then considered reached.

\section{Experiments in Simulation}

To validate the effectiveness of the proposed method, we conducted extensive experiments in simulated environments and evaluated the model's performance. Additionally, we performed comparative experiments to verify the effectiveness of CeRLP with compared methods.

\subsection{Experiment Setup}

The model is tested in the Gazebo simulation environment with communication handled through the Robot Operating System (ROS)\cite{quigley2009ros}. Key experimental parameters are as follows:
velocity $\omega_{\text{max}}=\pi/2 \text{ rad/s}$, acceleration $a_{v, \text{max}}=3 \text{ m/s}^{\text{2}}$, $a_{\omega, \text{max}}=3 \text{ rad/s}^{\text{2}}$.
$L_{\text{front}}=0.21 \text{ m}$, $L_{\text{rear}}=0.21 \text{ m}$, ${W}=0.5 \text{ m}$.

\begin{figure}[!t]
	\centering
	\includegraphics[width=\columnwidth]{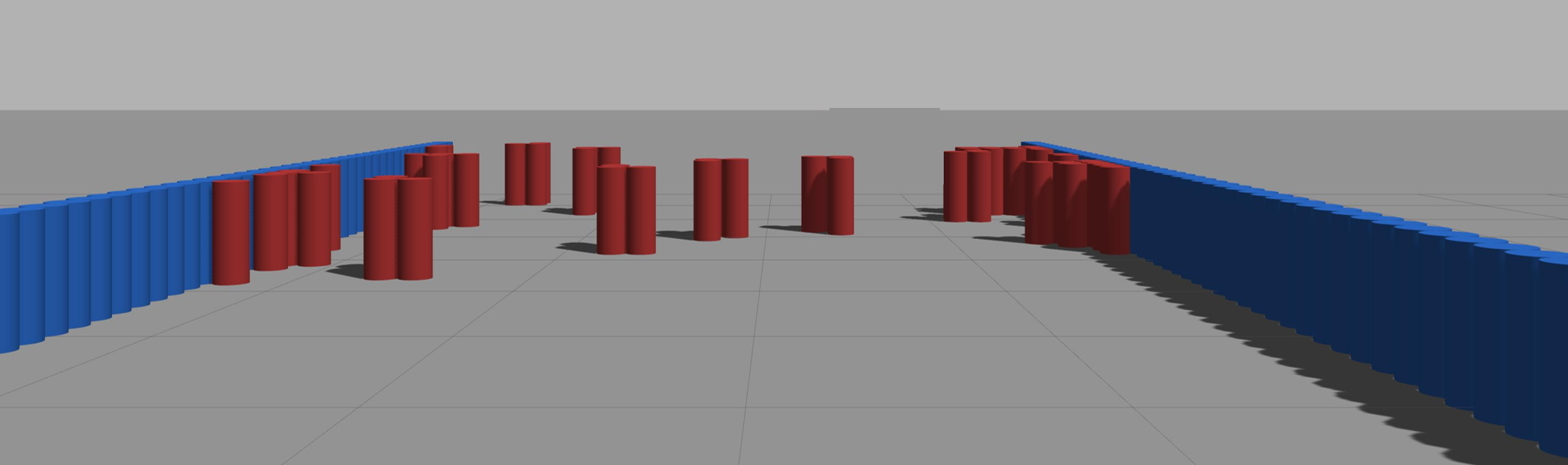}\\
	\caption{ Example of test environment in simulation.
	}\label{fig:simulation env}
\end{figure}

To evaluate the navigation performance of our model, we tested our method on the Benchmark for Autonomous Robot Navigation (BARN) dataset\cite{9292572} in Gazebo simulation. We randomly selected 100 different environments, with each environment tested twice. In the experimental scenario, the robot started at a fixed starting point and ended at a fixed goal. The Gazebo scene included randomly placed dense cylindrical obstacles. An example of a test environment is shown in Fig. \ref{fig:simulation env}.

We tested our approach with various compared methods to measure effectiveness: Dynamic Window Approach (DWA)\cite{580977}, Fast Dynamic Window Approach (Fast-DWA)\cite{10673780}, Elastic Bands (E-Band)\cite{291936}, and End-to-End (E2E)\cite{9975161}. We also compared CeRLP with the baseline method DRL-DCLP\cite{10900448}. E2E was trained using the test environments. 
In the simulation environment, all comparative methods took 2D LiDAR as input, whereas our method used a combination of laserscans generated from the front camera and the 2D LiDAR. Specifically, the camera generated laserscans covering a 75-degree range in front of the robot, while the LiDAR provided laserscans covering a 285-degree range on the left, right, and rear of the robot. The experimental platform was a Jackal four-wheeled differential-drive robot. The camera was a Flea3 camera, and the 2D LiDAR was a HOKUYO UST 10. 

To better compare different methods, we adopted the following evaluation metrics:

\textbf{Metric:} Following the BARN Challenge\cite{10673780}, this metric comprehensively considers task completion status and time efficiency:
\begin{equation}
\text{Metric} =\frac{1}{N}\sum_{\text{i}=1}^{N}{ S_{\text{i}} \cdot \frac{T_{\text{opt}, \text{i}}}{\text{clip}(T_{\text{act}, \text{i}}, 2T_{\text{opt}, \text{i}}, 8T_{\text{opt}, \text{i}})}},
\end{equation}
where $S_{\text{i}} \in \{0,1\}$ indicates whether the $i$-th test is successful, $T_{\text{opt}}$ represents the optimal path time, and $T_{\text{act}}$ denotes the actual execution time.

The optimal time is calculated from the path length and maximum velocity $v_{\text{max},\text{i}}$ in the $i$-th test, where $L_{\text{path}, \text{i}}$ is provided by the BARN dataset based on Dijkstra's search from the given start to goal position:
\begin{equation}
T_{\text{opt}, \text{i}} = \frac{L_{\text{path}, \text{i}}}{v_{\text{max},\text{i}}}.
\end{equation}

\textbf{Success Rate (SR):} This metric measures the percentage of tasks successfully completed across all test environments:
\begin{equation}
\text{SR} = \frac{1}{N}\sum_{\text{i}=1}^{N} S_{\text{i}},
\end{equation}
where $S_{\text{i}} \in \{0,1\}$ indicates whether the $i$-th test is successful.

\textbf{Collision Rate (CR):} This metric evaluates the collision behavior of the robot:
\begin{equation}
\text{CR} = \frac{1}{N}\sum_{\text{i}=1}^{N} C_{\text{i}},
\end{equation}
where $C_{\text{i}} \in \{0,1\}$ indicates whether a collision occurs in the $i$-th test. 

\textbf{Timeout Rate (TR):} This metric indicates whether the robot exceeds the time limit:
\begin{equation}
\text{TR} = \frac{1}{N}\sum_{\text{i}=1}^{N} O_{\text{i}},
\end{equation}
where $O_{\text{i}} \in \{0,1\}$ indicates whether the $i$-th test times out, measuring the algorithm's time efficiency.
\subsection{Experiment Result Analysis}

We evaluated the navigation performance under two maximum velocity settings: $0.5\,\mathrm{m/s}$ and $1.0\,\mathrm{m/s}$. The results demonstrated that our proposed CeRLP achieved superior performance compared to traditional methods and other learning-based baselines, exhibiting robustness comparable to the baseline method DRL-DCLP.

\begin{figure}[!t]
\centering
\subfloat[]{
		\includegraphics[width=0.3\columnwidth]{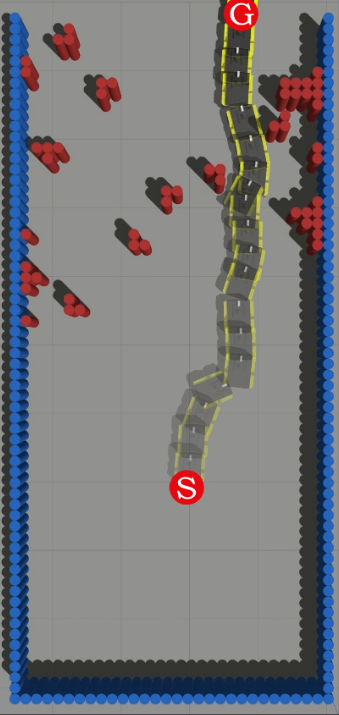}}
\subfloat[]{
		\includegraphics[width=0.3\columnwidth]{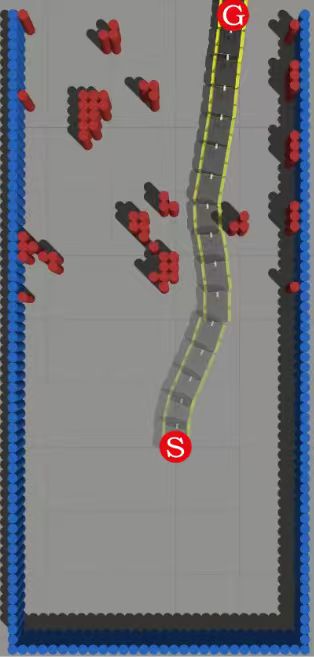}}
\subfloat[]{
		\includegraphics[width=0.3\columnwidth]{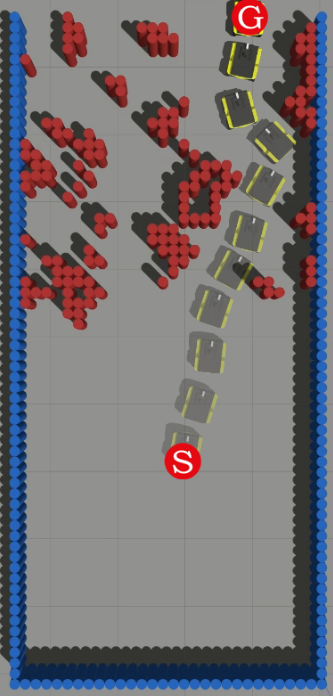}}
\caption{The visualization of CeRLP trajectories in test BARN environments. In (a) and (b), the maximum linear velocity was $0.5 \text{ m/s}$. In (c), the maximum linear velocity was $1 \text{ m/s}$.}
\label{fig_barn}
\end{figure}

\textbf{Quantitative Comparison:}
The quantitative results of the comparative experiments in the BARN simulation environments are shown in Table \ref{tab:comparison}. CeRLP outperforms DWA, Fast-DWA, E-Band, and E2E across both velocity settings. 
Specifically, at a velocity of $0.5\,\mathrm{m/s}$, CeRLP achieves a SR of $78.0\%$, surpassing the best-performing traditional planner E-Band by $13.0\%$. In terms of the comprehensive navigation metric, CeRLP scores $0.3790$, demonstrating its efficiency advantage. When the difficulty increases with a maximum velocity of $1.0\,\mathrm{m/s}$, CeRLP continues to outperform most of the compared methods. While the SR of DWA and Fast-DWA drops to below $45\%$, CeRLP maintains a robust SR of $70.0\%$. 
Notably, CeRLP achieves the lowest TO among all methods, with only $3.0\%$ at $0.5\,\mathrm{m/s}$ and $0.0\%$ at $1.0\,\mathrm{m/s}$. This indicates that our method efficiently finds feasible paths in dense clutter without getting stuck, whereas traditional methods often fail to escape local minima, leading to high timeout rates. The visualizations of simulation experiments are shown in Fig. \ref{fig_barn}. In the figure, the starting point is marked with a red "S," and the goal is marked with a red "G." This demonstrates the navigation and obstacle avoidance capabilities of our RL-based policy, which navigates through randomly placed dense cylindrical obstacles.

Although CeRLP outperforms all compared methods, it is second to DRL-DCLP. This is an expected result, as DRL-DCLP utilizes ground-truth 2D LiDAR data with a $360^\circ$ FOV and precise distance measurements. In contrast, CeRLP relies on a monocular camera for forward obstacle perception with a limited FOV and inherent depth estimation noise. 
The performance gap can be attributed to the errors introduced during the visual-to-scan conversion process. The depth estimation noise and the resolution loss during projection create a modality gap between the virtual scan and the real LiDAR scan. We analyze the specific impact of these visual errors in Section VI-B. 
Nevertheless, despite these sensory limitations, CeRLP still demonstrates navigation capabilities that are no less effective than those of the compared methods.

\begin{table*}[htbp]
\centering
\caption{Comparison of Different Methods using BARN Dataset under Two Velocity Settings}
\label{tab:comparison}
\begin{tabular}{@{}lccccccccccl@{}}
\toprule
\multirow{3}{*}{Method} & \multicolumn{4}{c}{$v_{\text{max}}$: 0.5 m/s} & \multicolumn{4}{c}{$v_{\text{max}}$: 1.0 m/s} & \multirow{3}{*}{\makecell{DRL-\\based}} & \multirow{3}{*}{\makecell{Use Global\\Path}} & \multirow{3}{*}{\makecell{Input\\Modality}} \\
\cmidrule(lr){2-5} \cmidrule(lr){6-9}
 & Metric ($\uparrow$) & SR ($\uparrow$)& CR ($\downarrow$)& TO ($\downarrow$)& Metric ($\uparrow$)& SR ($\uparrow$)& CR($\downarrow$) & TO ($\downarrow$)& & & \\
\midrule
DWA & 0.2354 & 55.0\% & 9.0\% & 36.0\% & 0.207 & 43.0\% & 26.0\% & 31.0\% & No & Yes & LiDAR \\
Fast-DWA & 0.2471 & 55.6\% & 11.1\% & 33.3\% & 0.2095 & 44.0\% & 23.0\% & 33.0\% & No & Yes & LiDAR \\
E-Band & 0.3244 & 65.0\% & \textbf{8.0\%} & 27.0\% & 0.2250 & 45.0\% & \textbf{7.0\%} & 48.0\% & No & Yes & LiDAR \\
E2E & 0.2520 & 53.0\% & 42.0\% & \underline{5.0\%} & 0.3439 & 69.0\% & 27.0\% & \underline{4.0\%} & Yes & No & LiDAR \\
DRL-DCLP & \textbf{0.4067} & \textbf{83.0\%} & \underline{9.0\%} & 8.0\% & \textbf{0.3561} & \textbf{73.0\%} & \underline{13.0\%} & 14.0\% & Yes & No & LiDAR \\
Ours & \underline{0.3790} & \underline{78.0\%} & 19.0\% & \textbf{3.0\%} & \underline{0.3544} & \underline{70.0\%} & 30.0\% & \textbf{0.0\%} & Yes & No & \textbf{RGB} \\
\bottomrule
\end{tabular}
\begin{flushleft}
\centering
\footnotesize The Best Result is in Bold, and the Second Best is Underlined.
\end{flushleft}
\end{table*}

\section{Experiments in the Real World}

\begin{table*}[t]
\centering
\caption{Robot Configurations for Real-world Cross-Embodiment Experiments}
\label{tab:robot_configs}
\renewcommand{\arraystretch}{1.2}
\begin{tabular}{@{}llccccclc@{}}
\toprule
\multirow{2}{*}{\textbf{Config ID}} & \multirow{2}{*}{\textbf{Platform Type}} & \multicolumn{3}{c}{\textbf{Physical Dimensions (m)}} & \multicolumn{2}{c}{\textbf{Camera Positions (m)}} & \multirow{2}{*}{\textbf{Camera Setup}} & \multirow{2}{*}{\textbf{$v_{\text{max}}$ (m/s)}} \\
\cmidrule(lr){3-5} \cmidrule(lr){6-7}
 &  & $L_{\text{front}}$ & $L_{\text{rear}}$ & $W$ & Height ($z_{\text{cam}}$) & Offset ($x_{\text{cam}}$) &  &  \\
\midrule
\textbf{DMR1} (Base) & Diff-Drive & 0.20 & 0.20 & 0.40 & 0.42 & +0.03 & 3 $\times$ WHEELTEC C100 & 0.5 \\
\textbf{DMR2} (Long) & Diff-Drive & 0.15 & 0.45 & 0.40 & 0.42 & +0.03 & 3 $\times$ WHEELTEC C100 & 0.5 \\
\textbf{DMR3} (Long) & Diff-Drive & 0.40 & 0.20 & 0.40 & 0.42 & +0.03 & 3 $\times$ WHEELTEC C100 & 0.5 \\
\textbf{DMR4} (Wide) & Diff-Drive & 0.20 & 0.20 & 1.00 & 0.42 & +0.03 & 3 $\times$ WHEELTEC C100 & 0.5 \\
\textbf{DMR5} (High) & Diff-Drive & 0.18 & 0.20 & 0.40 & 0.52 & -0.13 & 3 $\times$ WHEELTEC C100 & 0.5 \\
\midrule
\textbf{DMR6} (Low) & Diff-Drive & 0.20 & 0.20 & 0.40 & 0.20 & +0.15 & 1 $\times$ Orbbec Femto Bolt & 0.5 \\
\textbf{DMR7} (Quad) & Quadruped & 0.35 & 0.35 & 0.30 & 0.60 & +0.00 & 1 $\times$ WHEELTEC C100  & 0.6 \\
\bottomrule
\end{tabular}
\end{table*}

In this section, we first introduced the robot configurations and experiment setups for the real-world experiments. Next, we compared and analyzed the results of the Depth Estimation Scale Correction. Subsequently, we conducted real-world ablation studies to validate the effectiveness of our proposed method across varying physical dimensions, and camera types, camera positions, and dynamic obstacles. Finally, we evaluated the performance of VLN  with different robot sizes.

\begin{figure*}[!t]
	\centering
	\includegraphics[width=0.8\textwidth]{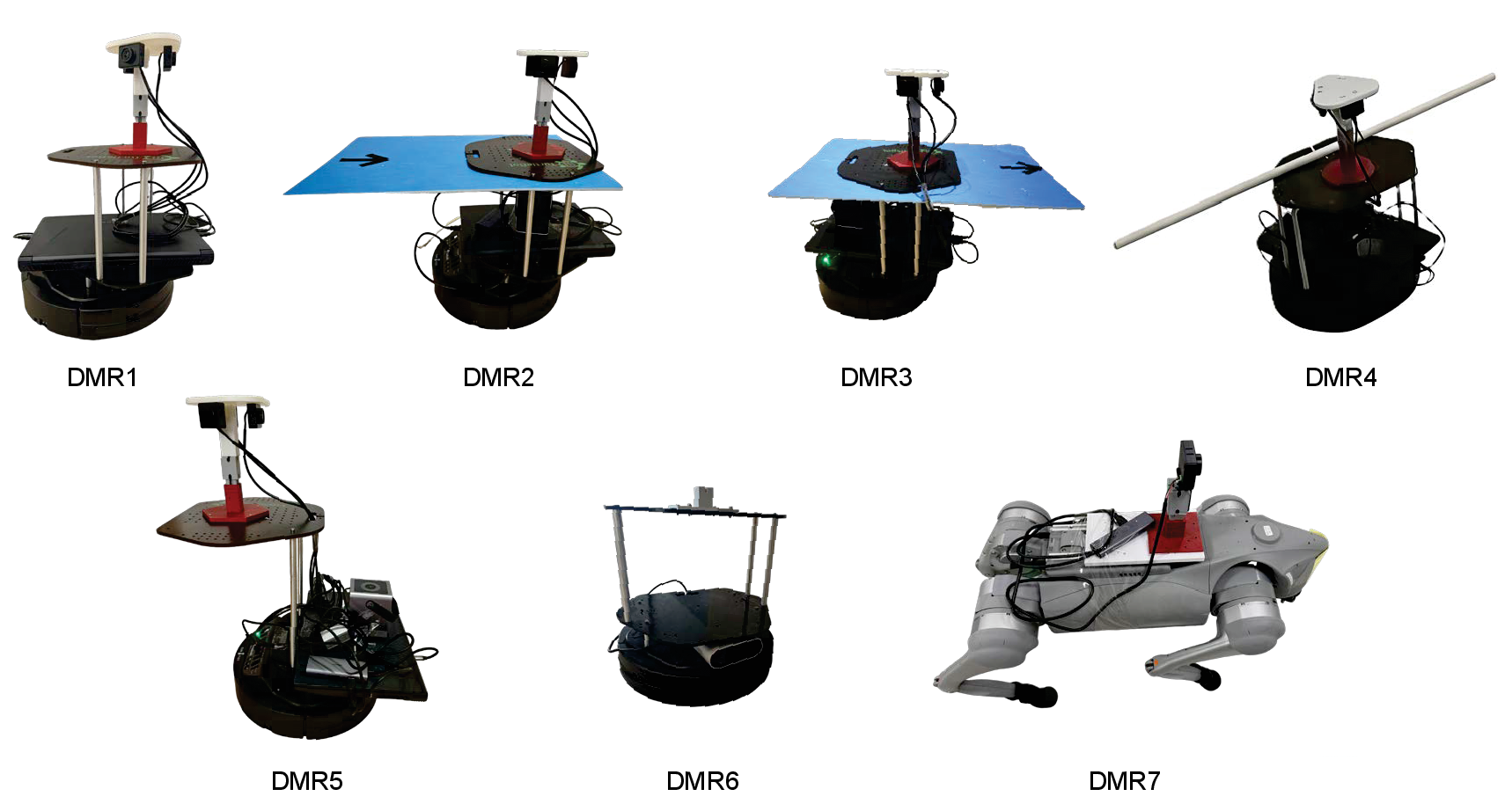}\\
	\caption{ Differential-Drive Mobile Robots (DMR). DMR1-DMR6 were wheeled robots based on the TurtleBot 2 chassis, varying in physical dimension, camera type, and camera position. DMR7 was a Unitree Go2 quadruped robot equipped with a WHEELTEC C100 camera.
	}\label{fig:robot_platform}
\end{figure*}

\subsection{Experiment Setup}

In the real-world experiments, we employed heterogeneous robotic platforms, including differential-drive wheeled robots and quadruped robots, to validate the zero-shot generalization capability of CeRLP.  We designed seven distinct differential-drive mobile robots (DMR1-DMR7) to cover a wide range of physical dimensions and sensor setups. The robot body dimensions were divided by the axis of the drive center, where the length of the front section was $L_{\text{front}}$, the length of the rear section was $L_{\text{rear}}$, and the width was $W$. The camera position was defined by the mounting height $z_{\text{cam}}$ and the longitudinal offset $x_{\text{cam}}$ relative to the drive center. $x_{\text{cam}}$ was positive towards the front and negative towards the rear. $z_{\text{cam}}$ was the camera's height above the ground.
The experimental platforms were detailed in Table \ref{tab:robot_configs}. The pictures of all tested platforms are shown in Fig. \ref{fig:robot_platform}.

\begin{itemize}
    \item Wheeled Platforms (DMR1-DMR6): These configurations were built on the Kobuki base TurtleBot 2. To achieve a comprehensive FOV, DMR1 through DMR5 were equipped with three WHEELTEC C100 cameras, which were combined to form a $360^\circ$ field of view. DMR6 deployed a single Orbbec Femto Bolt RGB-D camera to test performance under a narrow FOV constraint and utilized only the RGB image from the Orbbec Femto Bolt in the test. In DMR1-DMR6, all algorithms were executed on a laptop equipped with an NVIDIA RTX 4060 GPU.
    \item Quadruped Platform (DMR7): We utilized a Unitree Go2 robot equipped with a single WHEELTEC C100 camera. DMR7 was equipped with an NVIDIA Jetson Xavier NX for onboard computing. A statically placed Ultra Wide Band (UWB) device serves as the goal destination.
\end{itemize}

All algorithms used the Robot Operating System (ROS)\cite{quigley2009ros} for communication. The planning frequency is 10 Hz. In the real-world experiments, the robot employs odometry and IMU fusion for localization. When the robot is within 2 meters of the destination point, it is deemed to have reached the destination.
Key experimental parameters are as follows:
velocity $\omega_{\text{max}}=\pi/4 \text{ rad/s}$, acceleration $a_{v, \text{max}}=1 \text{ m/s}^{\text{2}}$, $a_{\omega, \text{max}}=\pi \text{ rad/s}^{\text{2}}$.

\begin{figure*}[htbp]
	\centering
	\includegraphics[width=0.7\textwidth]{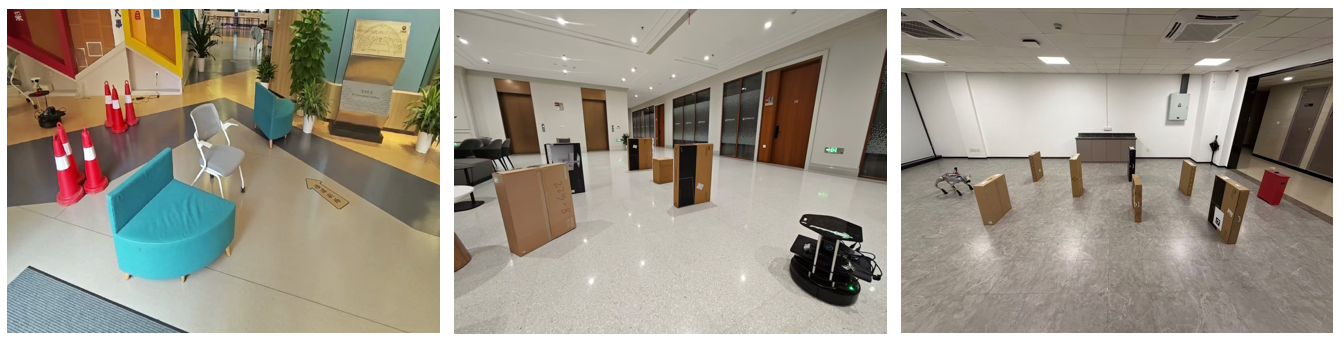}\\
	\caption{ Examples of test environments. From left to right: Env1, Env2, and Env3.
	}\label{fig:real env}
\end{figure*}

We conducted tests in three types of environments: (1) Env1: Featuring arbitrary obstacles such as roadblocks, sofas, and flower pots. (2) Env2: Containing structured obstacles like chairs and narrow passages formed by cardboard boxes. (3) Env3: Including densely packed cardboard boxes as obstacles, varying in shape, size, and color. The real-world environments are shown in Fig. \ref{fig:real env}.

\subsection{Results for Depth Estimation Scale Correction}

\begin{table}[!t]
  \centering
  \caption{Quantitative comparison of depth estimation methods}
  \label{tab:depth_comparison}
  \begin{tabular}{l
                  S[table-format=1. 4]
                  S[table-format=1.4]
                  S[table-format=3.2]}
    \toprule
    \textbf{Method} & {\textbf{MAE (m)}} & {\textbf{RMSE (m)}} & {\textbf{Time (ms)}} \\
    \midrule
    VGGT-1B\cite{11094896}         & 1.1459             & 1.2507              & 173.68 \\
    DAS-Outdoor\cite{37379163738604}    & 3.0189             & 3.8202              & 27.90  \\
    DAS-Indoor\cite{37379163738604}      & 4.1895             & 5.9287              & \textbf{26.85}  \\
    \midrule
    Ours+DAS        & 0.6947    & 1.2134      & 28.33  \\
    Ours+DAB        & 0.6038             &  1.1665             & 58.58  \\
    Ours+DAL        & \textbf{0.5866}    & \textbf{ 1.1214}              & 162.04 \\
    \bottomrule
  \end{tabular}
\end{table}

We evaluated the proposed depth estimation scale correction method by comparing the estimated depth results produced by different methods. The compared methods were: (1) our depth estimation scale correction using relative depth from Depth Anything V2\cite{37379163738604} (abbreviated as Ours+DAS, Ours+DAB, and Ours+DAL for the Small, Base, and Large versions respectively), and (2) metric depth estimation baselines, including VGGT-1B\cite{11094896} and Depth Anything V2 Small for metric depth estimation (DAS-Outdoor and DAS-Indoor).

In order to evaluate the results, we captured a real-world depth test dataset using a depth camera mounted on a TurtleBot2. A human operator teleoperated the robot through the test scenes while the camera recorded synchronized RGB images and ground-truth depth images. The depth camera ran at 25 Hz. After recording, we sampled the recorded stream at 1 Hz to form the final image dataset. The resulting depth test dataset contained 1000 RGB–depth image pairs. This depth test dataset was used only for evaluation, and there was no fine-tuning or training for all tested models.
All models were evaluated on a laptop with an RTX 4060 GPU. To meet real-time constraints for mobile deployment, every method was quantized to FP16 using TensorRT. All methods took an RGB image as input and produced a metric depth image as output.

Table \ref{tab:depth_comparison} shows the Mean Absolute Error (MAE) and Root Mean Square Error (RMSE) between predicted and ground-truth depth images, and the average processing time to get a depth image for each method. Our depth estimation scale correction methods and the Depth Anything variants achieve lower MAE and RMSE values. Among them, Ours+DAL achieves the lowest depth estimation error, while Ours+DAS maintains low depth estimation error with low processing time. 
In terms of processing time, all methods based on Depth Anything V2 Small (DAS) achieve real-time performance, with DAS-Outdoor, DAS-Indoor, and Ours+DAS exhibiting similar inference speeds. However, although DAS-Outdoor and DAS-Indoor can predict metric depth directly, they still exhibit errors in our experiments due to variations in camera type and camera positions. VGGT-1B produces metric depth with relatively low error but requires the longest processing time.

\subsection{Cross-Embodiment Navigation}

\begin{figure*}[!t] 
\centering
\subfloat[DMR1]{
		\includegraphics[width=0.24\textwidth]{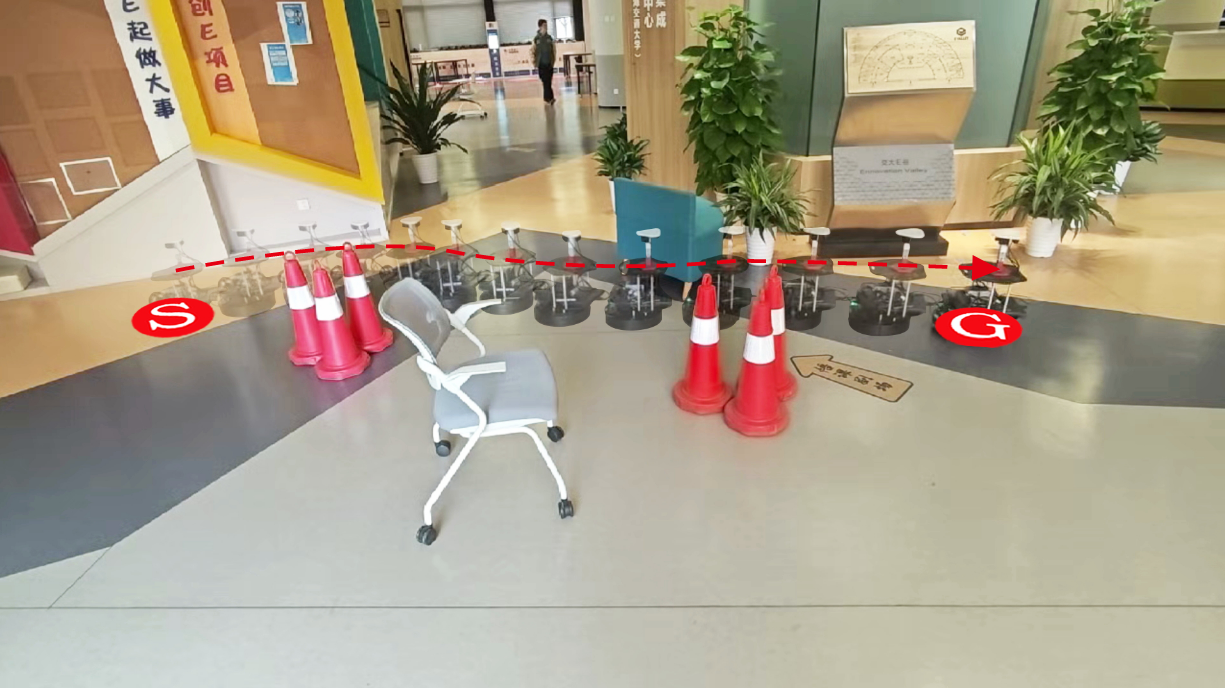}}
\hfill 
\subfloat[DMR2]{
		\includegraphics[width=0.24\textwidth]{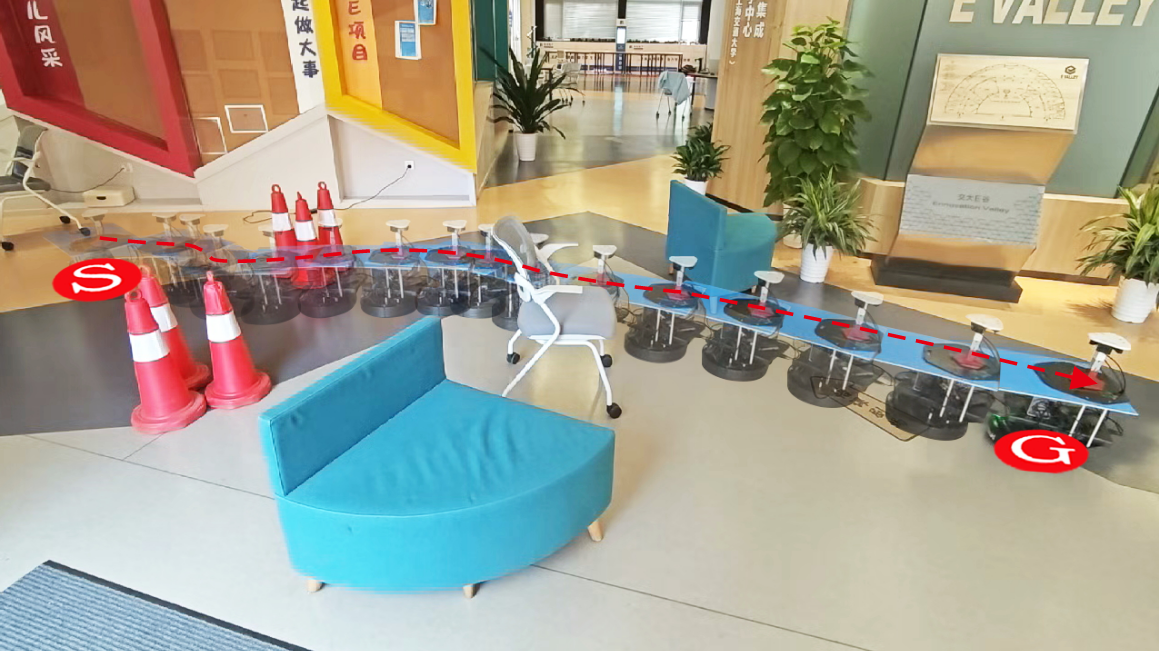}}
\hfill
\subfloat[DMR3]{
		\includegraphics[width=0.24\textwidth]{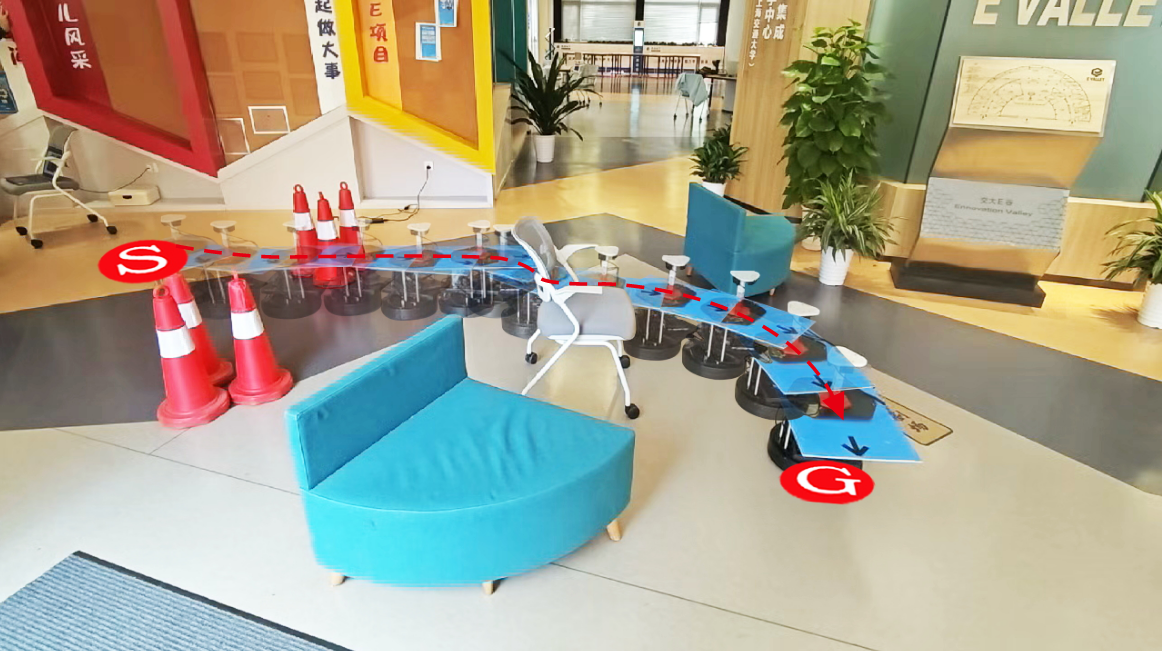}}
\hfill
\subfloat[DMR4]{
		\includegraphics[width=0.24\textwidth]{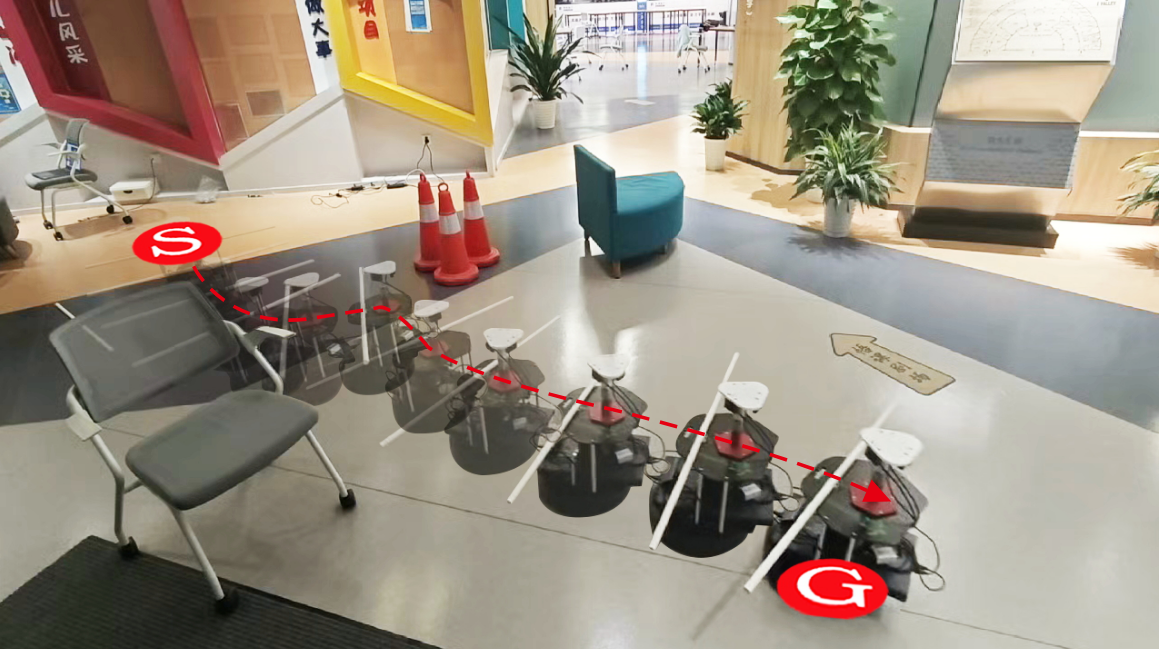}}
\caption{Real-world experiments for different physical dimensions.}
\label{fig_size}
\end{figure*}

\begin{figure*}[!t]
\centering
\subfloat[DMR1]{
		\includegraphics[width=0.24\textwidth]{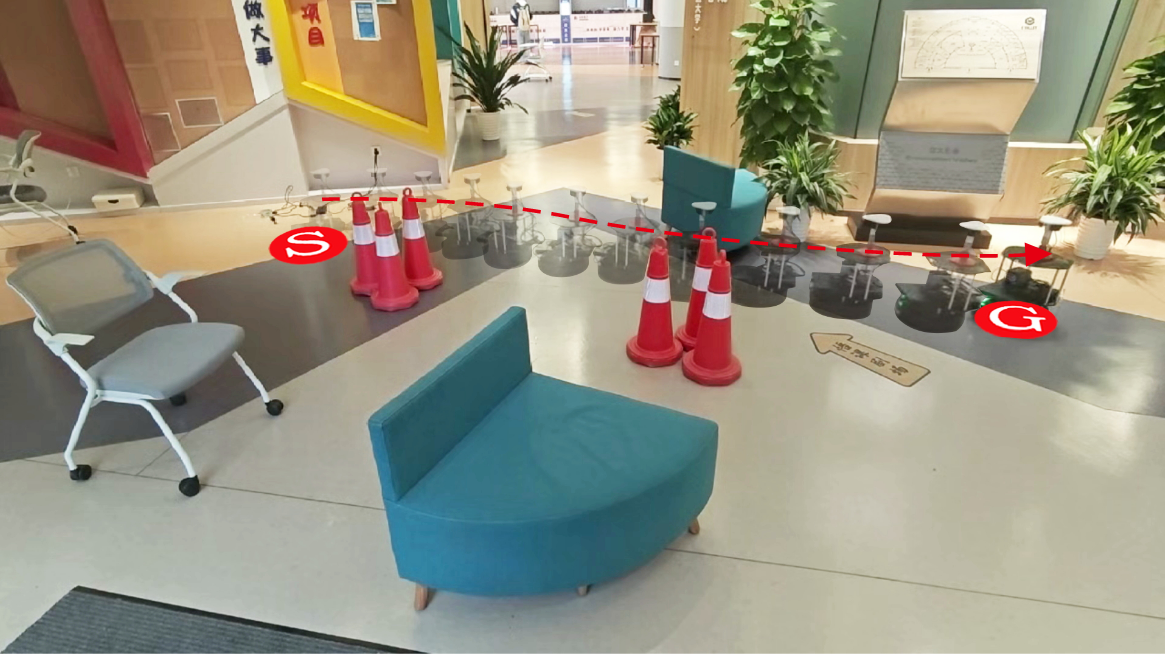}}
\hfill 
\subfloat[DMR5]{
		\includegraphics[width=0.24\textwidth]{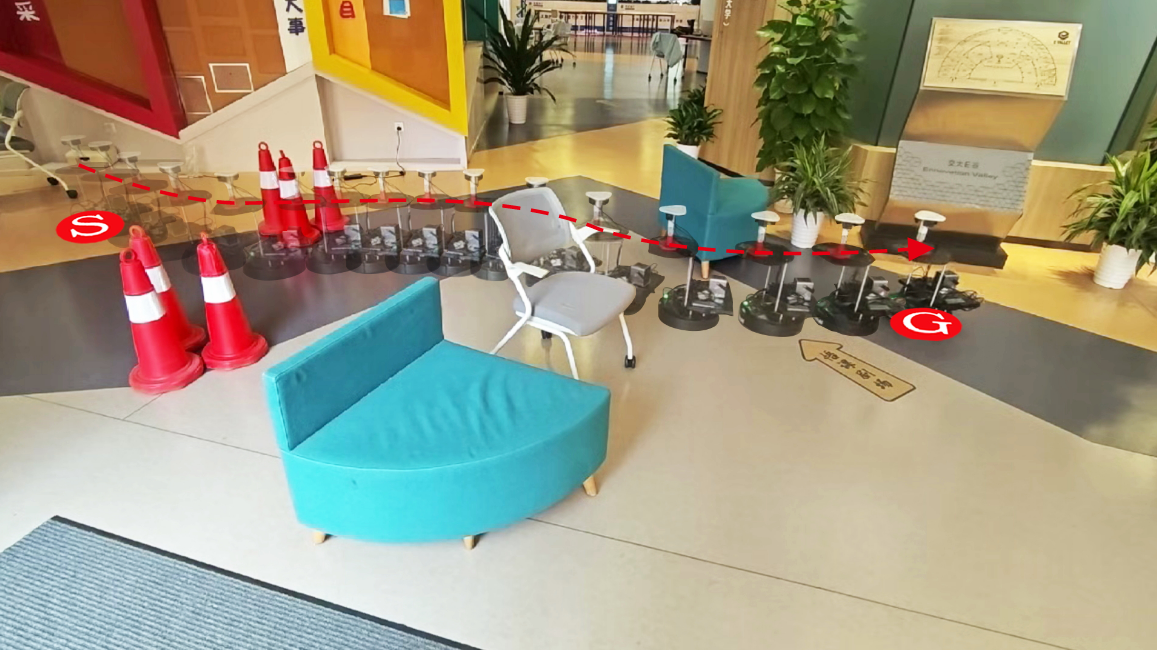}}
\hfill 
\subfloat[DMR5 with pedestrians]{
		\includegraphics[width=0.24\textwidth]{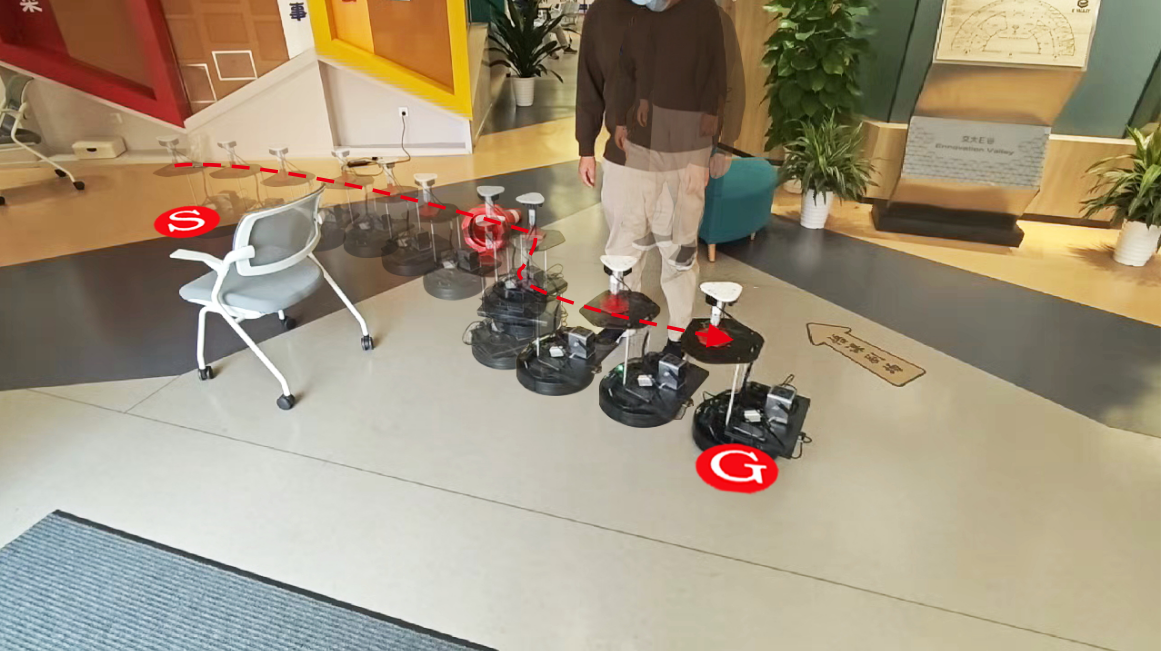}}
\hfill 
\subfloat[DMR6]{
		\includegraphics[width=0.24\textwidth]{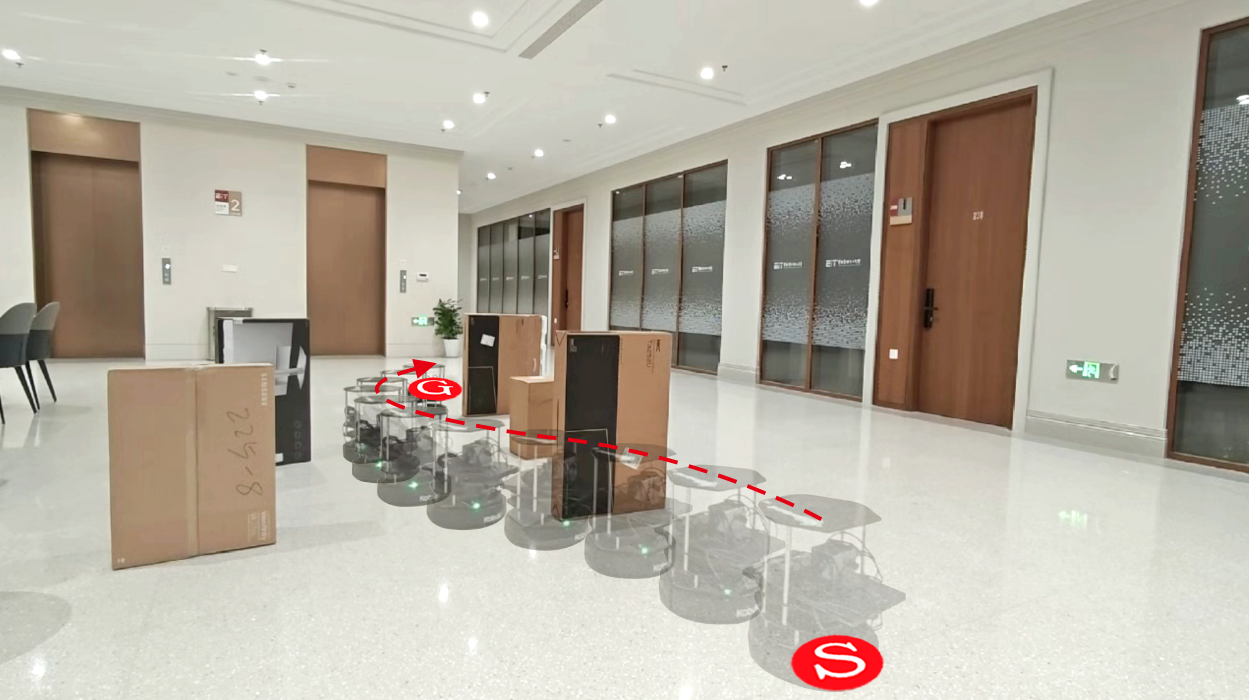}}
\caption{Real-world experiments for different camera positions.}
\label{fig_pos}
\end{figure*}

\begin{figure*}[!t] 
\centering
\subfloat[DMR1 with pedestrians]{
		\includegraphics[width=0.24\textwidth]{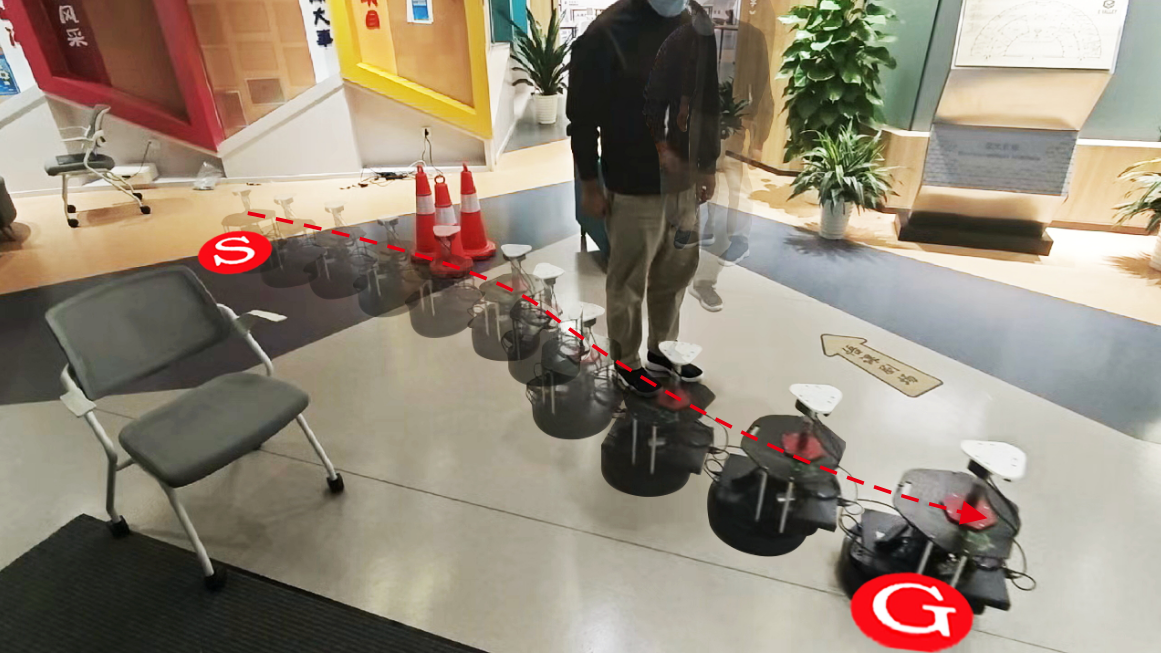}}
\hfill 
\subfloat[DMR2 with pedestrians]{
		\includegraphics[width=0.24\textwidth]{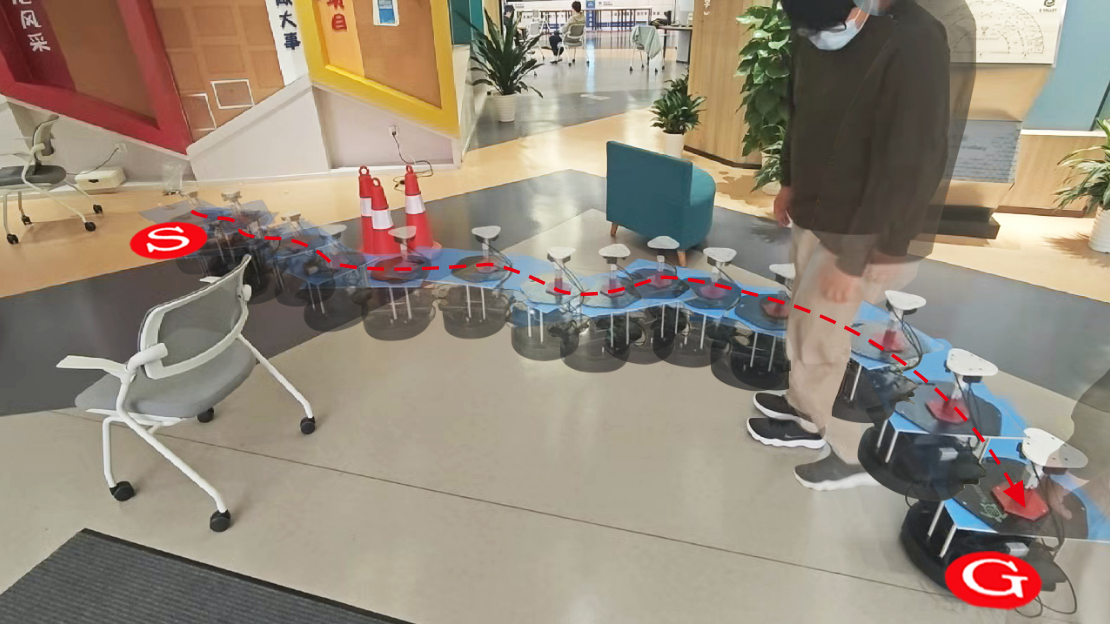}}
\hfill 
\subfloat[DMR3 with pedestrians]{
		\includegraphics[width=0.24\textwidth]{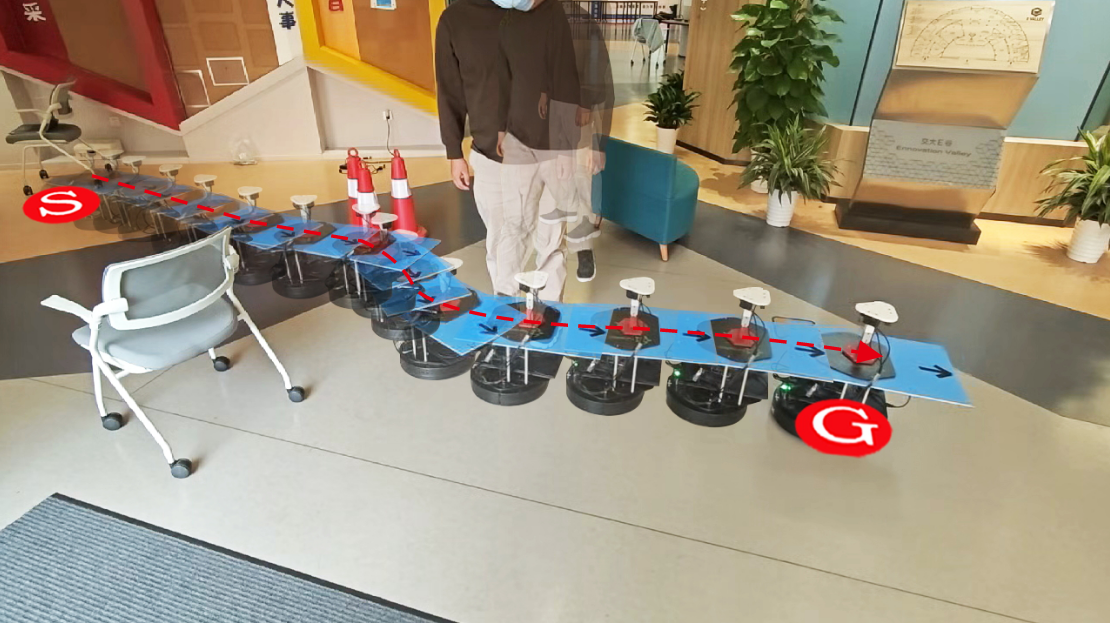}}
\hfill 
\subfloat[DMR7 with pedestrians]{
		\includegraphics[width=0.24\textwidth]{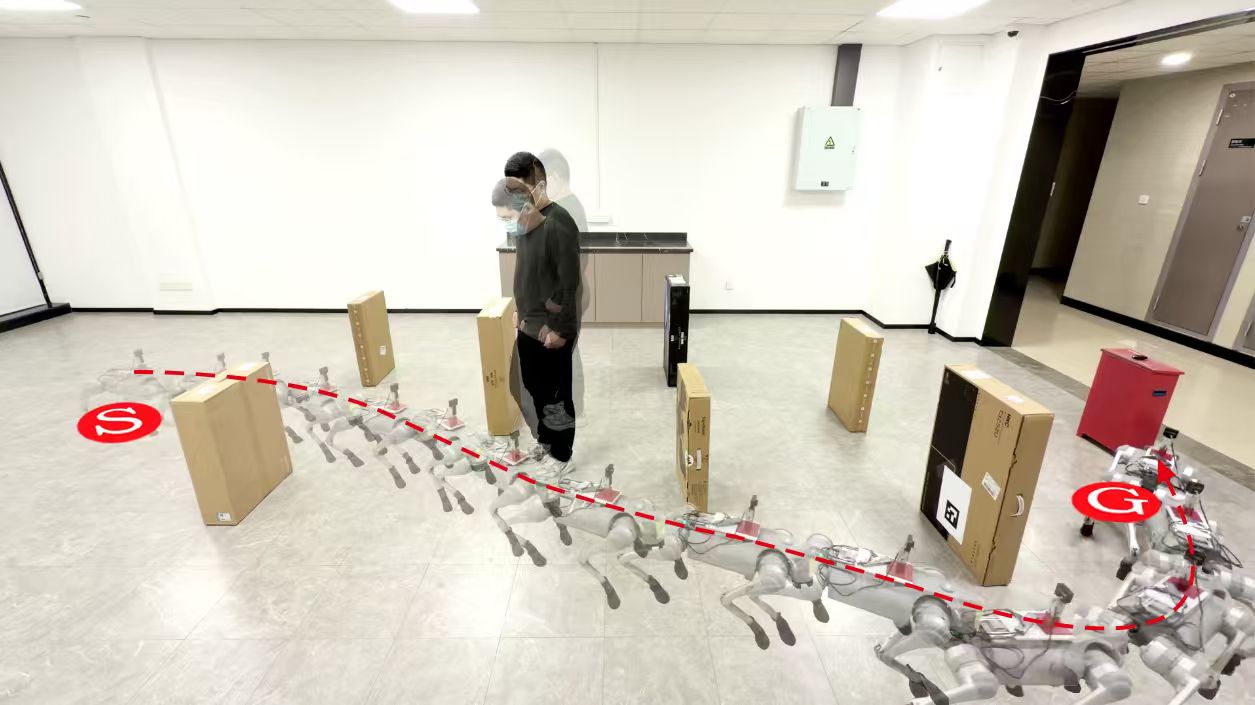}}
\caption{Real-world experiments for different physical dimensions with pedestrians.}
\label{fig_pedestrians}
\end{figure*}

\begin{figure}[!t]
	\centering
	\includegraphics[width=\columnwidth]{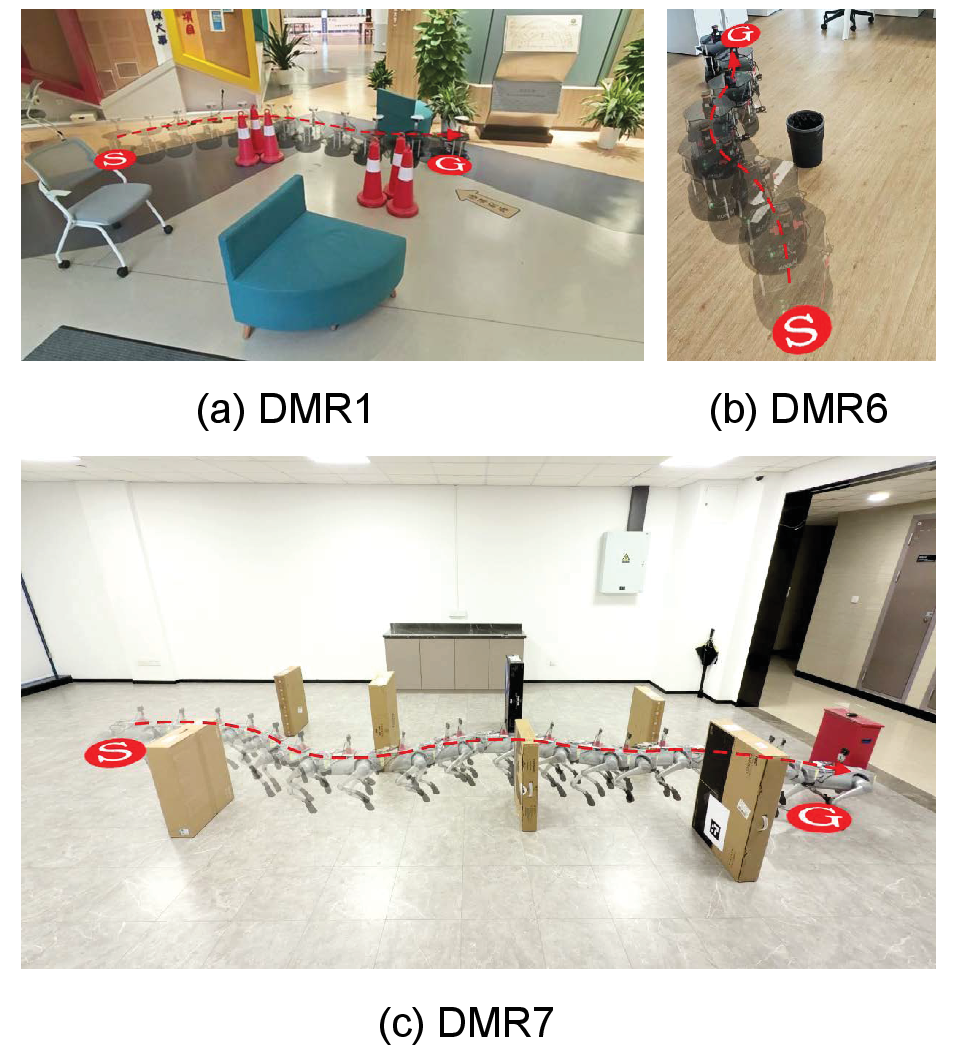}\\
	\caption{ Real-world experiments for different camera types.
	}\label{fig_type}
\end{figure}

In this subsection, we conducted ablation studies to verify the navigation performance of our method across different physical dimensions, camera types, and camera positions. In the following figures, the starting point was marked with a red "S", and the goal was marked with a red "G".

\textbf{Robustness to Physical Dimensions:}
We first evaluated the adaptability of CeRLP to changes in the robot's physical footprint using the $360^\circ$ camera setup. We compared four configurations:

\noindent \textbf{(a)} Standard Size (DMR1): A base footprint ($L_{\text{front}}=0.20\text{m}$, $L_{\text{rear}}=0.20\text{m}$).

\noindent \textbf{(b)} Long Rear (DMR2): An asymmetric configuration with an extended rear length ($L_{\text{front}}=0.15\text{m}$, $L_{\text{rear}}=0.45\text{m}$), testing the ability to avoid collisions during turns.

\noindent \textbf{(c)} Long Front (DMR3): An asymmetric configuration with an extended front length ($L_{\text{front}}=0.40\text{m}$, $L_{\text{rear}}=0.20\text{m}$), testing the ability to avoid collisions during turns.

\noindent \textbf{(d)} Extra Wide (DMR4): A configuration with a width of $1.0\text{m}$, testing the navigation performance of the robot when loaded with wide objects.

The experimental trajectories are shown in Fig. \ref{fig_size}. CeRLP successfully generated collision-free paths for all configurations. Notably, for the wide robot (DMR4), the planner correctly navigated across gaps between obstacles.

\textbf{Robustness to Camera Positions:}
Then, we investigated the impact of camera positions, which drastically altered the visual perspective and ground plane projection. We compared:

\noindent \textbf{(a)} Standard Position (DMR1): Camera at $z_{\text{cam}} = 0.42\text{m}$.

\noindent \textbf{(b)} High \& Back Position (DMR5): Camera mounted higher ($0.52\text{m}$) and shifted backward ($-0.13\text{m}$), creating a higher perspective near the robot front.

\noindent \textbf{(c)} High \& Back Position (DMR5) with pedestrians.

\noindent \textbf{(d)} Low \& Front Position (DMR6): Camera mounted lower ($0.20\text{m}$) and shifted forward ($+0.15\text{m}$), making the ground plane appear larger in the image.

The results in Fig. \ref{fig_pos} indicated that our height-adaptive projection successfully mitigated the perspective distortion. Specifically, in the low-mount scenario (DMR6), the system correctly filtered out the ground texture that baseline methods often misclassified as obstacles, ensuring safe navigation across varying extrinsic configurations.

\begin{figure*}[!t]
	\centering
	\includegraphics[width=\textwidth]{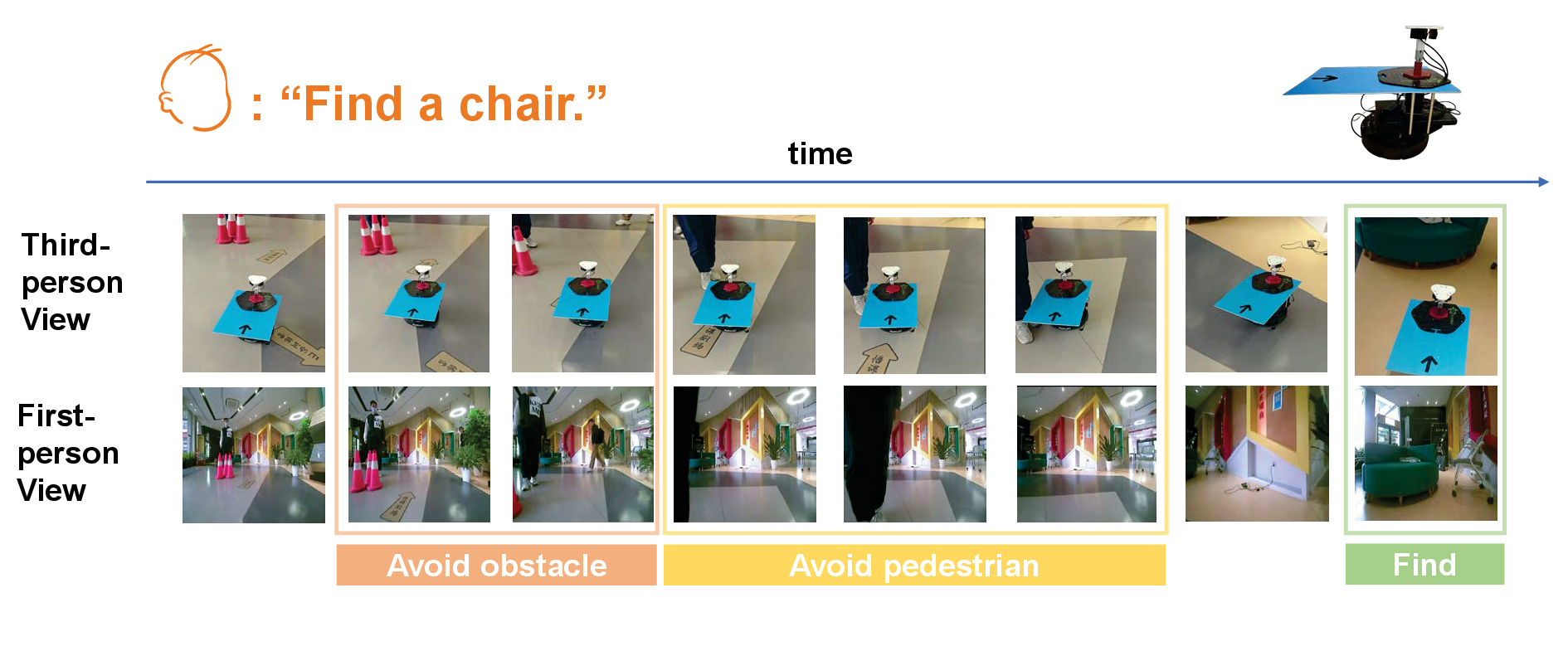}\\
	\includegraphics[width=\textwidth]{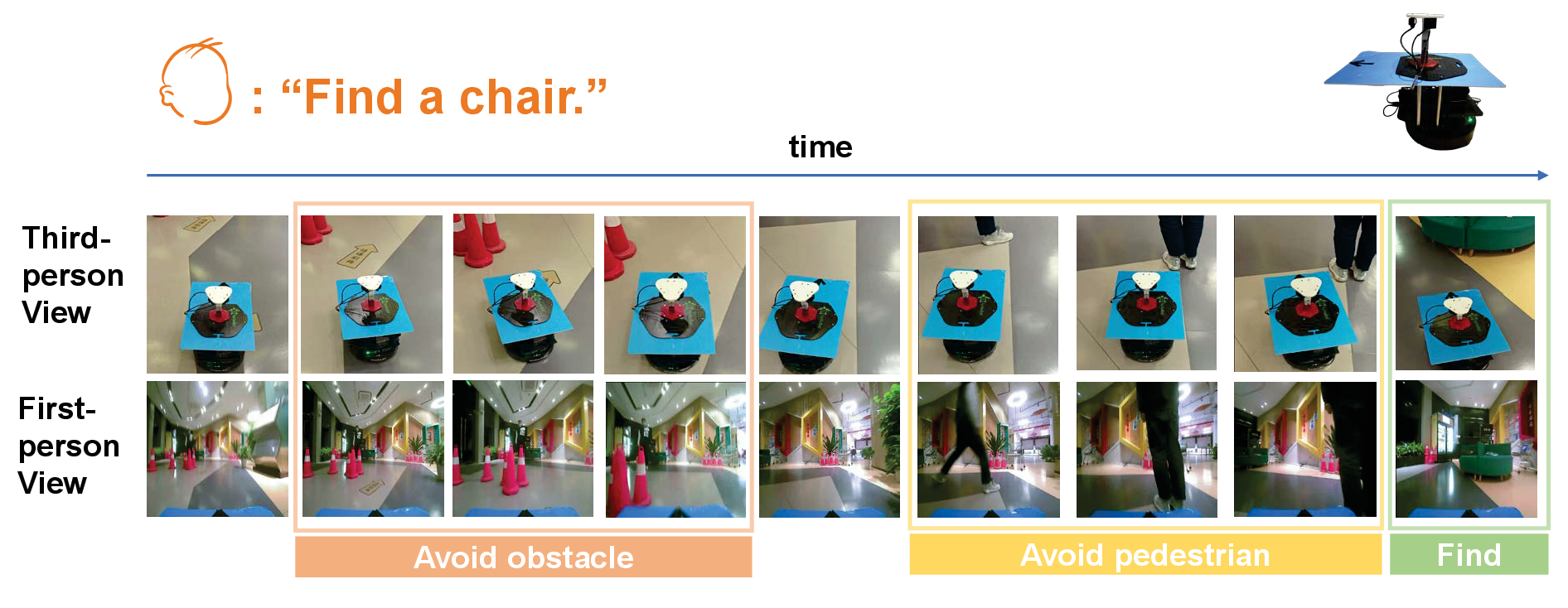}\\
	\caption{ Obstacle avoidance performance in VLN experiments. The images include both first-person and third-person views as time progresses. Test configurations are DMR2 and DMR3. The robot must find targets according to language instructions, with the VLN providing high-level commands. Concurrently, the robot should avoid obstacles within the scene and pedestrians who suddenly appear. The test language instruction is "Find a chair".
	}\label{fig: vln_obstacle_avoidance}
\end{figure*}

\textbf{Navigation with Pedestrians:}
Finally, we conducted navigation experiments to evaluate safety in dynamic environments. In these scenarios, in addition to static obstacles, pedestrians suddenly intruded into the robot's path, requiring the CeRLP to perform real-time reactive avoidance. We compared:

\noindent \textbf{(a)} Standard Size (DMR1): The baseline symmetric footprint ($L_{\text{front}}=0.20\text{m}, L_{\text{rear}}=0.20\text{m}$) on the wheeled chassis.

\noindent \textbf{(b)} Long Rear (DMR2): An asymmetric configuration with an extended rear ($L_{\text{front}}=0.15\text{m}, L_{\text{rear}}=0.45\text{m}$). This tested the ability to account for tail swing collision risks during sudden evasive maneuvers.

\noindent \textbf{(c)} Long Front (DMR3): An asymmetric configuration with an extended front ($L_{\text{front}}=0.40\text{m}, L_{\text{rear}}=0.20\text{m}$). This tested the ability to maintain safety margins for protruding frontal parts.

\noindent \textbf{(d)} Quadruped Robot (DMR7): A quadruped robot configuration ($L_{\text{front}}=0.35\text{m}, L_{\text{rear}}=0.35\text{m}$). This verified the model's obstacle avoidance capability on a completely different locomotion substrate.

The qualitative results are presented in Fig. \ref{fig_pedestrians}. It could be observed that CeRLP successfully generated safe trajectories for all configurations. Regardless of the robot size, camera height, or platform type, CeRLP correctly perceived the dynamic pedestrian as a high-risk obstacle and executed timely avoidance actions without colliding with surrounding static obstacles.

\textbf{Robustness to Camera Types:}
Next, we evaluated the system's robustness across different sensor types and FOV constraints. We compared:

\noindent \textbf{(a)} Wide View (DMR1): Using three synchronized C100 cameras to form a $360^\circ$  field of view. 

\noindent \textbf{(b)} Narrow View (DMR6): Using a single Orbbec camera with a limited forward-facing FOV.

\noindent \textbf{(c)} Quadruped Robot (DMR7): Using the single C100 camera of the Unitree Go2, which introduced the camera shake.

As shown in Fig. \ref{fig_type}, while the DMR1 allowed for smoother movement, the single-camera setups also successfully completed the navigation tasks. This demonstrated that our Visual-to-Scan module could effectively aggregate partial observations into a unified description, even with limited FOV or unstable camera motion.

\subsection{VLN Experiments with Different Robot Sizes}

We conducted Vision-Language Navigation (VLN) experiments to evaluate the obstacle avoidance capability of CeRLP in VLN tasks. We tested two different physical dimensions, DMR2 and DMR3. The test environments contained various static obstacles, including traffic cones and flowerpots, as well as a suddenly appearing pedestrian.

In these VLN tasks, the robot had to successfully avoid obstacles while navigating toward the specified language target object. As shown in  Fig. \ref{fig: vln_obstacle_avoidance}, CeRLP successfully avoided obstacles and found the target across different robot size configurations. This demonstrated that CeRLP had the ability to enhance the safety of robots as the lower-level planner for VLN. In our experiments, we achieved performance comparable to LiDAR sensors using only RGB cameras. This approach was more cost-effective for real-world applications compared to systems relying on LiDAR or depth cameras.

\section{Conclusion}

In this paper, we propose CeRLP, a general framework for visual navigation designed to effectively achieve cross-embodiment navigation across heterogeneous robots and camera configurations. We first introduce an offline scale calibration method to recover metric scale information from monocular depth estimation, enabling precise scale correction. Subsequently, we propose the Visual-to-Scan Abstraction module, which aggregates the corrected depth images into a unified virtual 2D laser scan. To address the variations in scan generation caused by diverse camera positions, we employ Height-Adaptive Obstacle Filtering to adaptively filter ground points and select optimal projection heights. We conduct extensive experiments in simulation, demonstrating that CeRLP outperforms comparative methods across various performance metrics. Furthermore, extensive real-world experiments verify the effectiveness of CeRLP in physical scenarios, while ablation studies confirm the contribution of individual modules and the robustness of the system under different configurations.

In future research, we plan to further explore the universality of CeRLP as a foundational baseline, including its integration with other dimension-configurable planners to validate safety across different strategies and the adaptability of CeRLP. Additionally, as this work demonstrates CeRLP's capability to execute high-level Vision-Language Navigation (VLN) commands and enhance obstacle avoidance, we intend to further leverage CeRLP to improve the safety and robustness of VLN agents in complex environments.


\bibliographystyle{IEEEtran}
\bibliography{main}

\end{document}